%% file: paper.tex
\documentclass{article}

\pdfoutput=1

\usepackage[preprint,nonatbib]{neurips_2021}
\usepackage[numbers]{natbib}
\input{math_commands.tex}

\usepackage{makecell}
\usepackage{microtype}
\usepackage{graphicx}
\usepackage{booktabs} 
\usepackage{amsfonts}       
\usepackage{nicefrac}       
\usepackage{microtype}      
\usepackage{graphicx}
\usepackage{subcaption}
\usepackage{amsmath}
\usepackage{amsthm}
\usepackage{amssymb}
\usepackage{mathtools}
\usepackage{xcolor}
\usepackage{csquotes}
\usepackage{algorithm}
\usepackage{algorithmicx}
\usepackage{algpseudocode}
\usepackage{wrapfig}
\usepackage{float}
\usepackage{tikz}

\usetikzlibrary{bayesnet}
\usetikzlibrary{arrows}

\usepackage{multibib}
\newcites{appendix}{References}

\newcommand\bs[1]{\boldsymbol{#1}}
\newcommand\bv[1]{\mathbf{#1}}

\newcommand{\fedsparse}{\texttt{FedSparse}}
\newcommand{\fedavg}{\texttt{FedAvg}}
\newcommand{\feddrop}{\texttt{FedDrop}}
\newcommand{\fedlone}{\texttt{FedL1}}
\newcommand{\scaffold}{\texttt{SCAFFOLD}}

\title{An Expectation-Maximization Perspective on Federated Learning}

\author{%
Christos Louizos$^\ddagger$,\enskip
Matthias Reisser$^\ddagger$,\enskip
Joseph Soriaga$^\ddagger$,\enskip
Max Welling$^\dagger$\\
$^\ddagger$Qualcomm AI Research \thanks{Qualcomm AI Research is an initiative of Qualcomm Technologies, Inc. and/or its subsidiaries.}, \enskip $^\dagger$University of Amsterdam\\
\texttt{\{clouizos,mreisser,jsoriaga\}@qti.qualcomm.com},\enskip\texttt{m.welling@uva.nl}
}

\begin{document}

\maketitle

\begin{abstract}
  Federated learning describes the distributed training of models across multiple clients while keeping the data private on-device. In this work, we view the server-orchestrated federated learning process as a hierarchical latent variable model where the server provides the parameters of a prior distribution over the client-specific model parameters. We show that with simple Gaussian priors and a \emph{hard} version of the well known Expectation-Maximization (EM) algorithm, learning in such a model corresponds to \fedavg{}, the most popular algorithm for the federated learning setting. This perspective on \fedavg{} unifies several recent works in the field and opens up the possibility for extensions through different choices for the hierarchical model. Based on this view, we further propose a variant of the hierarchical model that employs prior distributions to promote sparsity. By similarly using the hard-EM algorithm for learning, we obtain \fedsparse{}, a procedure that can learn sparse neural networks in the federated learning setting. \fedsparse{} reduces communication costs from client to server and vice-versa, as well as the computational costs for inference with the sparsified network – both of which are of great practical importance in federated learning.
\end{abstract}

\input{sections/intro.tex}
\input{sections/method.tex}
\input{sections/related.tex}
\input{sections/experiments.tex}

\input{sections/conclusion.tex}

\bibliography{paper}
\bibliographystyle{plain}

\appendix
\input{sections/appendix}
\end{document}

%% file: math_commands.tex
\usepackage{amsmath,amsfonts,bm}

\def\eqref#1{equation~\ref{#1}}

\def\1{\bm{1}}

\DeclareMathAlphabet{\mathsfit}{\encodingdefault}{\sfdefault}{m}{sl}
\SetMathAlphabet{\mathsfit}{bold}{\encodingdefault}{\sfdefault}{bx}{n}

\DeclareMathOperator*{\argmax}{arg\,max}

%% file: sections/intro.tex
\section{Introduction}
Smart devices have become ubiquitous in today's world and are generating large amounts of potentially sensitive data. Traditionally, such data is transmitted and stored in a central location for training machine learning models. Such methods rightly raise privacy concerns, and we seek the means for training powerful models, such as neural networks, without the need to transmit potentially sensitive data. To this end, Federated Learning (FL)~\citep{mcmahan2016communication} has been proposed to train global machine learning models without the need for participating devices to transmit their data to the server. The Federated Averaging (\fedavg{})~\citep{mcmahan2016communication} algorithm communicates the parameters of the machine learning model instead of the data itself, which is a more private means of communication.

In this work, we adopt a perspective on server-orchestrated federated learning through the lens of a simple hierarchical latent variable model; the server provides the parameters of a prior distribution over the client-specific neural network parameters, which are subsequently used to ``explain'' the client-specific dataset. A visualization of this graphical model can be seen in Figure~\ref{fig:fedavg_hlvm}. We show that when the server provides a Gaussian prior over these parameters, learning with the \emph{hard} version of the Expectation-Maximization (EM) algorithm will yield the \fedavg{} procedure. This view of \fedavg{} has several interesting consequences, as it connects several recent works in federated learning and bridges \fedavg{} with meta-learning. 

Furthermore, this view of federated learning provides a good basis for developing several extensions to the vanilla \fedavg{} algorithm by appropriately changing the hierarchical model. Through this perspective, we develop \fedsparse{} by extending the graphical model to the one in Figure~\ref{fig:fedsparse_hlvm}. \fedsparse{} allows for learning sparse neural network models at the client and server via a careful choice of the priors within the hierarchical model. 
\input{graphical_models/both}

Sparse neural networks are appealing for the FL scenario, especially in the ``cross-device'' setting~\citep{kairouz2019advances}, as they can tackle several practical challenges. In particular, communicating model updates over multiple rounds across many devices can incur significant communication costs. Communication via the public internet infrastructure and mobile networks is potentially slow and not for free. Equally important, training (and inference) takes place on-device and is therefore restricted by the edge devices' hardware constraints on memory, speed and heat dissipation capabilities. Therefore, \fedsparse{} provides models that tackle both of the challenges mentioned above since they can simultaneously reduce the overall communication and computation at the client devices. Addressing these challenges is an important step towards building practical FL systems, as also discussed in~\cite{kairouz2019advances}. Empirically, we see that \fedsparse{} provides better communication-accuracy trade-offs compared to, \fedavg{}, prior works that similarly target joint reductions of computation and communication costs ~\citep{caldas2018expanding}, as well as another baseline that uses $L_1$ regularization at the client level (for pruning) and standard parameter averaging at the server level. 

%% file: graphical_models/both.tex
\begin{figure*}[ht]
\centering
\begin{minipage}[]{0.4\linewidth}
\vspace{-3em}
\centering
 \tikz{
 \node[obs] (Ds) {$D_s$};%
 \node[latent,left=of Ds] (phis) {$\boldsymbol{\phi}_s$}; %
 \node[det, above=of phis] (w) {$\mathbf{w}$};%
 \plate [] {plate1} {(Ds)(phis)} {$S$ shards}; %
 \tikzset{plate caption/.style={caption, node distance=0, inner sep=-3pt,
        below right=5pt and 5pt of #1.south,text height=1.2em,text depth=0.3em}};
 \plate [] {plate2} {(w)} {server};%
 \edge {phis} {Ds} 
 \edge {w} {phis}
 }
 \caption{The simple hierarchical model for server-orchestrated federated learning. With a Gaussian prior for the local parameters $\bs{\phi}_s$ centered at the server parameters $\bv{w}$ and hard-EM for learning we obtain \fedavg{}.}
 \label{fig:fedavg_hlvm}
\end{minipage}\hfill%
\begin{minipage}[ht]{0.55\linewidth}
\centering
\tikz{
 \node[obs] (Ds) {$D_s$};%
 \node[latent,left=of Ds] (phis) {$\boldsymbol{\phi}_s$}; %
 \node[det,above=of phis] (w) {$\mathbf{w}$};%
 \node[latent,left=of phis] (zs) {$\mathbf{z}_s$};%
 \node[det,above=of zs] (theta) {$\boldsymbol{\theta}$};%
 \plate [] {plate1} {(Ds)(phis)(zs)} {$S$ shards}; %
 \tikzset{plate caption/.style={caption, node distance=0, inner sep=-3pt,
        below right=5pt and -5pt of #1.south,text height=1.2em,text depth=0.3em}};
 \plate [] {plate2} {(w)(theta)} {server};%
 \edge {phis} {Ds} 
 \edge {w} {phis}
 \edge {zs} {phis}
 \edge {theta} {zs}
 }
 \caption{Modifying the hierarchical model to allow for sparsity in the local parameters $\bs{\phi}_s$ with the spike and slab distribution~\citep{mitchell1988bayesian}. The server parameter $\bs{\theta}$ governs a Bernoulli prior distribution over the local variables $\bv{z}_s$, the value of which determine whether a parameter in $\bs{\phi}_s$ is active or not. When a parameter is active, the prior is a Gaussian centered at the server parameters $\bv{w}$. By using hard-EM on this variant of the hierarchical model, we obtain \fedsparse{}.}
 \label{fig:fedsparse_hlvm}
\end{minipage}
\end{figure*}

%% file: sections/method.tex
\section{A hierarchical model interpretation of federated learning}
\label{sec:fedavg_as_em}
The server-orchestrated variant of federated learning is mainly realized via the \fedavg{}~\citep{mcmahan2016communication} algorithm, which is a simple iterative procedure consisting of four simple steps. At the beginning of each round $t$, the server communicates the model parameters, let them be $\bv{w}$, to a subset of the devices. The devices then proceed to optimize $\bv{w}$, \emph{e.g.}, via stochastic gradient descent, on their respective dataset via a given loss function
\begin{align}
    \mathcal{L}_s(\mathcal{D}_s, \bv{w}) := \frac{1}{N_s}\sum_{i=1}^{N_s} L(\mathcal{D}_{si}, \bv{w})
\end{align}
where $s$ indexes the device, $\mathcal{D}_s$ corresponds to the dataset at device $s$ and $N_s$ corresponds to its size. After a specific amount of epochs of optimization on $\mathcal{L}_s$ is performed, denoted as $E$, the devices communicate the current state of their parameters, let it be $\bs{\phi}_s$, to the server. The server then performs an update to its own model by simply averaging the client specific parameters 
$
    \bv{w}_t = \frac{1}{S}\sum_s \bs{\phi}_s.
$

\subsection{\fedavg{} as learning in a hierarchical model}
We now ask the following question; does the \fedavg{} algorithm correspond to learning on a specific model?
Let us consider the following objective function:
\begin{align}
    \argmax_\bv{w} \sum_{s=1}^S \log p(\mathcal{D}_s | \bv{w}),
\end{align}
where $\mathcal{D}_s$ corresponds to the shard specific dataset that has $N_s$ datapoints, $p(\mathcal{D}_s | \bv{w})$ corresponds to the likelihood of $\mathcal{D}_s$ under the server parameters $\bv{w}$. Now consider decomposing each of the shard specific likelihoods as follows:
\begin{align}
    p(\mathcal{D}_s | \bv{w}) = \int p(\mathcal{D}_s | \bs{\phi}_s) p(\bs{\phi}_s | \bv{w})\mathrm{d}\bs{\phi}_s, \quad p(\bs{\phi}_s | \bv{w}) =\frac{1}{Z} \exp\left(-\frac{\lambda}{2} \|\bs{\phi}_s - \bv{w}\|^2\right)\label{eq:prior_fedavg}, 
\end{align}
where we introduced the auxiliary latent variables $\bs{\phi}_s$, which are the parameters of the local model at shard $s$. The server parameters $\bv{w}$ act as the center for the Gaussian prior at Eq.~\ref{eq:prior_fedavg} over the shard specific parameters with the precision $\lambda$ acting as a regularization strength that prevents $\bs{\phi}_s$ from moving too far from $\bv{w}$. By putting everything together, the total objective will be
\begin{align}
    \mathcal{F}(\bv{w}) := \sum_s \log \int p(\mathcal{D}_s | \bs{\phi}_s) \frac{\exp\left(-\frac{\lambda}{2} \|\bs{\phi}_s - \bv{w}\|^2\right)}{Z} \mathrm{d}\bs{\phi}_s, \label{eq:total_obj_fedavg}
\end{align}
and the graphical model of this process can be seen at Figure~\ref{fig:fedavg_hlvm}. 

How can we then optimize Eq.~\ref{eq:total_obj_fedavg} in the presence of these latent variables $\bs{\phi}_s$? The traditional way to optimize such objectives is through Expectation-Maximization (EM). EM consists of two steps, the E-step where we form the posterior distribution over these latent variables
\begin{align}
p(\bs{\phi}_s | \mathcal{D}_s, \mathbf{w}) = \frac{p(\mathcal{D}_s | \bs{\phi}_s) p(\bs{\phi}_s | \bv{w})}{p(\mathcal{D}_s | \bv{w})},
\end{align}
and the M-step where we maximize the probability of $\mathcal{D}_s$ w.r.t. the parameters of the model $\bv{w}$ by marginalizing over this posterior
\begin{align}
    \argmax_\bv{w}\sum_s\mathbb{E}_{p(\bs{\phi}_s | \mathcal{D}_s, \mathbf{w}_{\text{old}})}[\log p(\mathcal{D}_s, \bs{\phi}_s | \bv{w})] = \argmax_\bv{w} \sum_s  \mathbb{E}_{p(\bs{\phi}_s | \mathcal{D}_s, \bv{w}_{\text{old}})}[\log p(\bs{\phi}_s| \bv{w})].
\end{align}
If we perform a single gradient step for $\bv{w}$ in the M-step, this procedure corresponds to doing gradient ascent on the original objective at Eq.~\ref{eq:total_obj_fedavg}, a fact we show in Appendix \ref{app:singleStepEMGradAscent}.

When posterior inference is intractable, hard-EM is usually employed as a simpler alternative. In this case we make ``hard'' assignment for the latent variables $\bs{\phi}_s$ in the E-step by approximating $p(\bs{\phi}_s | \mathcal{D}_s, \bv{w})$ with its most probable point, \emph{i.e.}
\begin{align}
    \bs{\phi}^*_s = \argmax_{\bs{\phi}_s} \frac{p(\mathcal{D}_s | \bs{\phi}_s) p(\bs{\phi}_s | \bv{w})}{p(\mathcal{D}_s | \bv{w})} = \argmax_{\bs{\phi}_s} \log p(\mathcal{D}_s | \bs{\phi}_s) + \log p(\bs{\phi}_s | \bv{w}).
\end{align}
This is usually easier to do as we can use techniques such as stochastic gradient ascent. Given these hard assignments, the M-step then corresponds to another simple maximization
\begin{align}
    \argmax_\bv{w} \frac{1}{S}\sum_s \log p(\bs{\phi}^*_s| \bv{w}).
\end{align}
As a result, performing hard-EM on the objective of Eq.~\ref{eq:total_obj_fedavg} corresponds to a block coordinate ascent type of algorithm on the following objective function
\begin{align}
    \argmax_{\bs{\phi}_{1:S}, \bv{w}} \sum_s (\log p(\mathcal{D}_s | \bs{\phi}_s) + \log p(\bs{\phi}_s | \bv{w})),\label{eq:hard_em}
\end{align}
where we alternate between optimizing $\bs{\phi}_{1:S}$ and $\bv{w}$ while keeping the other fixed.

How does this learning procedure correspond to \texttt{FedAvg}? By letting $\lambda \rightarrow 0$ in Eq.~\ref{eq:prior_fedavg} it is clear that the hard assignments in the E-step mimic the process of optimizing a local model on the data of each shard. In fact, even by optimizing the model locally with stochastic gradient ascent for a fixed number of iterations with a given learning rate, we implicitly assume a specific prior over the parameters; for linear regression, this prior is a Gaussian centered at the initial value of the parameters~\citep{santos1996equivalence} whereas for non-linear models it bounds the distance from the initial point. 
After obtaining $\bs{\phi}^*_s$ the M-step then corresponds to 
\begin{align}
    \argmax_{\bv{w}} \mathcal{L}_r := \sum_s -\frac{\lambda}{2}\|\bs{\phi}^*_s - \bv{w}\|^2 + C,
\end{align}
and we can easily find a closed form solution by setting the derivative of the objective w.r.t. $\bv{w}$ to zero and solving for $\bv{w}$:
\begin{align}
    \frac{\partial\mathcal{L}_r}{\partial \bv{w}} & = 0 \Rightarrow 
    \lambda \sum_s  \left(\bs{\phi}^*_s - \bv{w}\right) = 0 \Rightarrow 
     \bv{w} = \frac{1}{S}\sum_s \bs{\phi}^*_s.
\end{align}
It is easy to see that the optimal solution for $\bv{w}$ given $\bs{\phi}^*_{1:S}$ is the same as the one from \texttt{FedAvg}. Notice that for the ``cross-device'' settings of FL, the sum over all of the shards in the objective can be expensive. Nevertheless, it can easily be approximated by selecting a subset of the shards $B$, i.e., $\mathcal{L}_r \approx \frac{S}{B}\sum_{s\in B} -\frac{\lambda}{2}\|\bs{\phi}^*_s - \bv{w}\|^2 + C$. In this way, we only have to estimate $\bs{\phi}^*_s$ for the selected shards since the averaging for $\bv{w}$ involves only the selected shards.

Of course,~\texttt{FedAvg} does not optimize the local parameters $\bs{\phi}_s$ to convergence at each round, so one might wonder whether this correspondence is still valid. It turns out that the alternating procedure of EM corresponds to block coordinate ascent on a single objective function, the variational lower bound of the marginal log-likelihood~\citep{neal1998view} of a given model. More specifically for our setting, we can see that the EM iterations perform block coordinate ascent on: 
\begin{align}
    \sum_s \mathbb{E}_{q_{\bv{w}_s}(\bs{\phi}_s)}&\big[\log p(\mathcal{D}_s| \bs{\phi}_s)+\log p(\bs{\phi}_s | \mathbf{w})]+ H[q_{\bv{w}_s}(\bs{\phi}_s)]\label{eq:em_vlb}
\end{align}
to optimize $\bv{w}_{1:S}$ and $\bv{w}$, where $\bv{w}_s$ are the parameters of the variational approximation to the posterior distribution $p(\bs{\phi}_s | \mathcal{D}_s, \bv{w})$ and $H[q]$ corresponds to the entropy of the $q$ distribution. To obtain the hard-EM procedure, and thus~\texttt{FedAvg}, we can use a (numerically) deterministic distribution for $\bs{\phi}_s$, $q_{\bv{w}_s}(\bs{\phi}_s) := \mathcal{N}(\bv{w}_s, \epsilon\bv{I})$. This leads us to the same objective as in Eq.~\ref{eq:hard_em}, since the expectation concentrates on a single term and the entropy of $q_{\bv{w}_s}(\bs{\phi}_s)$ becomes a constant independent of the optimization. In this case, the optimized value for $\bs{\phi}_s$ after a fixed number of steps corresponds to the $\bv{w}_s$ of the variational approximation.

It is interesting to contrast recent literature under the lens of this procedure. Optimizing the same model with hard-EM but with a non-trivial $\lambda$ results in the same procedure that was proposed by~\cite{li2018federated}. Furthermore, using the difference of the local parameters to the global parameters as a ``gradient''~\citep{reddi2020adaptive} is equivalent to hard-EM on the same model, where in the M-step, instead of a closed-form update, we take a single gradient step and absorb the scaling $\lambda$ in the learning rate. In addition, this view makes precise the idea that \texttt{FedAvg} is a meta-learning algorithm~\citep{jiang2019improving}; the underlying hierarchical model it optimizes is similar to the ones used in meta-learning~\citep{grant2018recasting,chen2019modular}.

Besides connecting recent work in FL, this view also serves as a good basis for new algorithms for federated learning. 
The most straightforward way is to use an alternative prior in the hierarchical model, resulting in different local training and server-side updating behaviors. For example, one could use a Laplace prior, $p(\bs{\phi}_s|\bv{w}) = \text{Lap}(\bv{w}, \frac{1}{\lambda})$, which would result into the server selecting the median instead of averaging, or a mixture of Gaussians prior, $p(\bs{\phi}_s|\bv{w}_{1:K}, \frac{1}{\lambda}) = \frac{1}{K}\sum_k N(\bv{w}_k, \frac{1}{\lambda})$, which would result into training an ensemble of models at the server. We provide the details in Appendix~\ref{sec:alt_priors_fl}. In this work, we focus on tackling the communication and computational costs of FL, which is important and highly beneficial for practical applications of ``cross-device'' FL~\cite{kairouz2019advances}. For this reason, we replace the Gaussian prior with a sparsity inducing prior, namely the spike and slab prior~\cite{mitchell1988bayesian}. We describe the resulting algorithm, \fedsparse{}, in the next section.

\section{The \texttt{FedSparse} algorithm: sparsity in federated learning}
Encouraging sparsity in FL has two main advantages; the model becomes smaller, thus less resource-intensive to evaluate, and it cuts down on communication costs as the pruned parameters do not need to be communicated. The golden standard for sparsity in probabilistic models is the spike and slab~\citep{mitchell1988bayesian} prior. It is a mixture of two components, a delta spike at zero, $\delta(0)$, and a continuous distribution over the real line, \emph{i.e.} the slab. More specifically, by adopting a Gaussian slab for each local parameter $\bs{\phi}_{si}$ we have that
\begin{align}
    p(\bs{\phi}_{si}| \bs{\theta}_i, \bv{w}_i) = (1 - \bs{\theta}_i) \delta(0) + \bs{\theta}_i \mathcal{N}(\bs{\phi}_{si}| \bv{w}_i, 1/\lambda),
\end{align}
or equivalently as a hierarchical model
\begin{align}
    p(\bv{z}_{si}) = \text{Bern}(\bs{\theta}_i), \enskip
    & p(\bs{\phi}_{si}|\bv{z}_{si}=1, \bv{w}_i) = \mathcal{N}(\bs{\phi}_{si}| \bv{w}_i, 1/\lambda), \enskip
    p(\bs{\phi}_{si}|\bv{z}_{si}=0) = \delta(0) \\
    & p(\bs{\phi}_{si}| \bs{\theta}_i, \bv{w}_i) = \sum_{\bv{z}_{si}} p(\bv{z}_{si}|\bs{\theta}_i) p(\bs{\phi}_{si}|\bv{z}_{si}, \bv{w}_i),
\end{align}
where $\bv{z}_{si}$ plays the role of a ``gating'' variable that switches on or off the parameter $\bs{\phi}_{si}$. 
We thus modify the hierarchical model at Figure~\ref{fig:fedavg_hlvm} to use this new prior. The resulting hierarchical model can be seen in Figure~\ref{fig:fedsparse_hlvm}, where $\bv{w}$, $\bs{\theta}$ will be the server-side model weights and probabilities of the binary gates.

Following the \fedavg{} paradigm of simple point estimation, we will use hard-EM in order to optimize $\bv{w},\bs{\theta}$.
By using approximate distributions $q_{\bv{w}_s}(\bs{\phi}_s| \bv{z}_s)$, $q_{\bs{\pi}_s}(\bv{z}_s)$, the variational lower bound for this model becomes
\begin{align}
    \sum_s & \mathbb{E}_{q_{\bs{\pi}_s, \bv{w}_s}(\bv{z}_s, \bs{\phi}_s)}\big[\log p(\mathcal{D}_s| \bs{\phi}_s) + \log p(\bs{\phi}_s, \bv{z}_s| \mathbf{w}, \bs{\theta}) - \log q_{\bv{w}_s}(\bs{\phi}_s| \bv{z}_s)\big] + H[q_{\bs{\pi}_s}(\bv{z}_s)],\label{eq:init_elbo_kl_w}
\end{align}
which is to be optimized with respect to $\bv{w}_{1:S}, \bv{w}, \bs{\pi}_{1:S}, \bs{\theta}$.
For the shard specific weight distributions, as they are continuous, we will use $q_{\bv{w}_s}(\bs{\phi}_{si} | \bv{z}_{si} = 1) := \mathcal{N}(\bv{w}_{si}, \epsilon), q(\bs{\phi}_{si} | \bv{z}_{si} = 1) := \mathcal{N}(0, \epsilon)$ with $\epsilon \approx 0$ which will be, numerically speaking, deterministic. For the gating variables, as they are binary, we will use $q_{\bs{\pi}_{si}}(\bv{z}_{si}) := \text{Bern}(\bs{\pi}_{si})$ with $\bs{\pi}_{si}$ being the probability of activating local gate $\bv{z}_{si}$. In order to do hard-EM for the binary variables, we will remove the entropy term for the $q_{\bs{\pi}_s}(\bv{z}_s)$ from the aforementioned bound as this will encourage the approximate distribution to move towards the most probable value for $\bv{z}_s$. Furthermore, by relaxing the spike at zero to a Gaussian with precision $\lambda_2$, \emph{i.e.}, $p(\bs{\phi}_{si} | \bv{z}_{si} = 0) = \mathcal{N}(0, 1/\lambda_2)$, and by plugging in the appropriate expressions into Eq.~\ref{eq:init_elbo_kl_w} we can show that the local and global objectives will be
\begin{align}
    \mathcal{L}_s(\mathcal{D}_s, \bv{w}, \bs{\theta}, \bv{w}_s, \bs{\pi}_s) & := C + \mathbb{E}_{q_{\bs{\pi}_s}(\bv{z}_s)}\left[\sum_i^{N_s}L(\mathcal{D}_{si}, \bv{w}_s \odot \bv{z}_s)\right] - \frac{\lambda}{2} \sum_j\bs{\pi}_{sj}(\bv{w}_{sj} - \bv{w}_j)^2  \nonumber \\ &\quad - \lambda_0 \sum_j \bs{\pi}_{sj}  + \sum_j \left( \bs{\pi}_{sj}\log\bs{\theta}_j + (1 - \bs{\pi}_{sj}) \log(1 - \bs{\theta}_j)\right), \label{eq:local_lb_ss}\\
    \mathcal{L} & := \sum_{s=1}^{S}\mathcal{L}_s(\mathcal{D}_s, \bv{w}, \bs{\theta}, \bv{w}_s, \bs{\pi}_s)
\end{align}
respectively, where $\lambda_0 = \frac{1}{2}\log\frac{\lambda_2}{\lambda}$ and $C$ is a constant independent of the variables to be optimized. The derivation can be found in Appendix~\ref{sec:local_loss}. It is interesting to see that the final objective at each shard intuitively tries to find a trade-off between four things: 1) explaining the local dataset $\mathcal{D}_s$, 2) having the local weights close to the server weights (regulated by $\lambda$), 3) having the local gate probabilities close to the server probabilities and 4) reducing the local gate activation probabilities to prune away a parameter (regulated by $\lambda_0$). The latter is an $L_0$ regularization term, akin to the one proposed by~\cite{louizos2017learning}.

Now let us consider what happens at the server after the local shard, through some procedure, optimized $\bv{w}_s$ and $\bs{\pi}_s$. Since the server loss for $\bv{w}, \bs{\theta}$ is the sum of all local losses, the gradient for each of the parameters will be
\begin{align}
    \frac{\partial \mathcal{L}}{\partial \bv{w}} = \sum_s \lambda \bs{\pi}_s (\bv{w}_s - \bv{w}), \enskip
    \frac{\partial \mathcal{L}}{\partial \bs{\theta}} = \sum_s\left( \frac{\bs{\pi}_s}{\bs{\theta}} - \frac{1 - \bs{\pi}_s}{1 - \bs{\theta}}\right). \label{eq:fedspase_grad}
\end{align}
Setting these derivatives to zero, we see that the stationary points are
\begin{align}
    \bv{w} = \frac{1}{\sum_j \bs{\pi}_j}\sum_s \bs{\pi}_s\bv{w}_s, \qquad
    \bs{\theta} = \frac{1}{S}\sum_s \bs{\pi}_s \label{eq:fedsparse_avg}
\end{align}
\emph{i.e.}, a weighted average of the local weights and an average of the local probabilities of keeping these weights. Therefore, since the $\bs{\pi}_s$ are being optimized to be sparse through the $L_0$ penalty, the server probabilities $\bs{\theta}$ will also become small for the weights that are used by only a small fraction of the shards.
As a result, to obtain the final sparse architecture, we can prune the weights whose corresponding, learned, server inclusion probabilities $\bs{\theta}$ are less than a threshold, \emph{e.g.}, $0.1$. It should again be noted that the sums and averages of Eq.~\ref{eq:fedspase_grad},~\ref{eq:fedsparse_avg} respectively can be easily approximated by subsampling a small set of clients $S'$ from $S$. Therefore we do not have to consider all of the clients at each round, which would be prohibitive for ``cross-device'' FL.

\subsection{Reducing the communication cost}
The framework described so far allows us to learn a more efficient model. We now discuss how we can use it in order to cut down both download and upload communication costs during training.

\paragraph{Reducing client to server communication cost}
In order to reduce the client to server cost, we will communicate sparse samples from the local distributions instead of the distributions themselves; in this way, we do not have to communicate the zero values of the parameter vector, which leads to large savings.  More specifically, we can express the gradients and stationary points for the server weights and probabilities as follows 
\begin{align}
    \frac{\partial \mathcal{L}}{\partial \bv{w}} & = \sum_s \lambda \mathbb{E}_{q_{\bs{\pi}_s}(\bv{z}_s)}\left[\bv{z}_s(\bv{w}_s - \bv{w})\right], \quad
    \bv{w} = \mathbb{E}_{q_{\bs{\pi}_{1:S}}(\bv{z}_{1:S})}\left[\frac{1}{\sum_j \bv{z}_j}\sum_s \bv{z}_s\bv{w}_s\right], \\
    \frac{\partial \mathcal{L}}{\partial \bs{\theta}} & = \sum_s\mathbb{E}_{q_{\bs{\pi}_s}(\bv{z}_s)}\left[\frac{\bv{z}_s}{\bs{\theta}} - \frac{1 - \bv{z}_s}{1 - \bs{\theta}}\right],\quad
    \bs{\theta} = \frac{1}{S}\sum_s\mathbb{E}_{q_{\bs{\pi}_s}(\bv{z}_s)}\left[ \bv{z}_s\right].
\end{align}
As a result, we can then communicate from the client only the subset of the local weights $\hat{\bv{w}}_s$ that are non-zero in $\bv{z}_s \sim q_{\bs{\pi}_s}(\bv{z}_s), \quad \hat{\bv{w}}_s = \bv{w}_s \odot \bv{z}_s$, along with the $\bv{z}_s$. 
Having access to those samples, the server can then form 1-sample stochastic estimates of either the gradients or the stationary points for $\bv{w}, \bs{\theta}$. Notice that this is a way to reduce communication without adding bias in the gradients of the original objective. 
In case that we are willing to incur extra bias, future work can consider techniques such as quantization~\citep{amiri2020federated} and top-k gradient selection~\citep{lin2017deep} to reduce communication even further. 

\paragraph{Reducing the server to client communication cost}
The server needs to communicate to the clients the updated distributions at each round. Unfortunately, for simple unstructured pruning, this doubles the communication cost as for each weight $\bv{w}_i$ there is an associated $\bs{\theta}_i$ that needs to be sent to the client. To mitigate this effect, we will employ structured pruning, which introduces a single additional parameter for each group of weights. For groups of moderate sizes, \emph{e.g.}, the set of weights of a given convolutional filter, the extra overhead is small. 
We can also take the communication cost reductions one step further if we allow for some bias in the optimization procedure; we can prune the global model during training after every round and thus send to each of the clients only the subset of the model that has survived. Notice that this is easy to do and does not require any data at the server.  The inclusion probabilities $\bs{\theta}$ are available at the server, so we can remove the parameters that have $\bs{\theta}$ less than a threshold, \emph{e.g.} $0.1$. This can lead to large reductions in communication costs, especially once the model becomes sufficiently sparse.

\subsection{\fedsparse{} in practice}
\paragraph{Local optimization}
While optimizing for $\bv{w}_s$ locally is straightforward to do with gradient-based optimizers, $\bs{\pi}_s$ is more tricky, as the expectation over the binary variables $\bv{z}_s$ in Eq.~\ref{eq:local_lb_ss} is intractable to compute in closed form and using Monte-Carlo integration does not yield reparametrizable samples. To circumvent these issues, we rewrite the objective in an equivalent form and use the hard-concrete relaxation from~\citep{louizos2017learning}, which can allow for the straightforward application of gradient ascent. We provide the details in Appendix \ref{app:localgates}. When the client has to communicate to the server, we propose to form $\hat{\bv{w}}_s$ by sampling from the zero-temperature relaxation, which yields exact binary samples. Furthermore, at the beginning of each round, following the practice of \fedavg{}, the participating clients initialize their approximate posteriors to be equal to the priors that were communicated from the server. Empirically, we found that this resulted in better global model accuracy. 

\paragraph{Parameterization of the probabilities} There is evidence that such optimization-based pruning can be inferior to simple magnitude-based pruning~\citep{gale2019state}. We, therefore, take an approach that combines the two and reminisces the recent work of~\cite{azarian2020learned}. We parameterize the probabilities $\bs{\theta}, \bs{\pi}_s$ as a function of the model weights and magnitude-based thresholds that regulate how active a parameter can be:
\begin{align}
    \theta_g := \sigma\left(\frac{\|\bv{w}_g\|_2 - \tau_g}{T}\right),  \enskip
    \pi_{sg} := \sigma\left(\frac{\|\bv{w}_{sg}\|_2 - \tau_{sg}}{T}\right),
\end{align}
where the subscript $g$ denotes the group, $\sigma(\cdot)$ is the sigmoid function, $\tau_g, \tau_{sg}$ are the global and client specific thresholds for a given group $g$ and $T$ is a temperature hyperparameter. Following~\cite{azarian2020learned} we also ``detach'' the gradient of the weights through $\bs{\theta}, \bs{\pi}_s$, to avoid decreasing the probabilities by just shrinking the weights. With this parametrization we lose the ability to get a closed form solution for the server thresholds, but nonetheless we can still perform gradient based optimization at the server by using the chain rule. For a positive threshold, we use a parametrization in terms of a softplus function, \emph{i.e.}, $\bs{\tau} = \log(1 + \exp(\bv{v}))$ where $\bv{v}$ is the learnable parameter. The \fedsparse{} algorithm is described in Alg.~\ref{alg:fedsparse},~\ref{alg:fedsparse2} in the Appendix. 

%% file: sections/related.tex
\section{Related work}
\texttt{FedProx}~\citep{li2018federated} adds a proximal term to the local objective at each shard, so that it prevents the local models from drifting too far from the global model, while still averaging the parameters at the server. In Section~\ref{sec:fedavg_as_em} we showed how such a procedure arises if we use a non-trivial precision for the Gaussian prior over the local parameters of the hierarchical model and apply hard-EM. Furthermore, \fedavg{} has been advocated to be a meta-learning algorithm in~\cite{jiang2019improving}; with our perspective, this claim is precise and shows that the underlying hierarchical model that \fedavg{} optimizes is the same as the models used in several meta-learning works~\citep{grant2018recasting,chen2019modular}. Furthermore, by performing a single gradient step for the M-step in the context of hard-EM applied at the model of Section~\ref{sec:fedavg_as_em}, we see that we arrive at a procedure that has been previously explored both in a meta-learning context with the Reptile algorithm~\citep{nichol2018first}, as well as the federated learning context with the ``generalized'' \fedavg{}~\citep{reddi2020adaptive}. One important difference between meta-learning and \fedavg{} is that the latter maximizes the sum, across shards, marginal-likelihood in order to update the globa; parameters, whereas meta-learning methods usually optimize the global parameters such that the finetuned model performs well on the local validation sets. Exploring such parameter estimation methods, as, \emph{e.g.}, in \cite{chen2019modular}, in the federated scenario and how these relate to existing approaches that merge meta-learning with federated learning, \emph{e.g.} \cite{fallah2020personalized}, is an interesting avenue for future work. Finally, our EM perspective also applies to optimization works that improve model performance via model replicas~\citep{zhang2019lookahead,pittorino2020entropic}.

Adopting a hierachical model perspective for federated learning is not new and has been explored previously in, \emph{e.g.}, \cite{yurochkin2019bayesian,yurochkin2019statistical,corinzia2019variational,al2020federated}. \cite{yurochkin2019bayesian,yurochkin2019statistical} consider the local parameters as ``given'' and then fit a prior model to uncover latent structure. This is only a part of our story; from our EM perspective, the \emph{prior affects the local optimization itself}, thus it allow us to, \emph{e.g.}, sparsify a neural network locally with \fedsparse{}. 
\cite{corinzia2019variational,al2020federated} are more similar to our work; they are both \fedavg{}-like procedures that can be viewed as a way to \emph{aggregate inferences} across clients in order to get a ``global'' posterior approximation (albeit with different procedures) to the parameters of the server model. This is different to our view of \fedavg{} as learning the parameters of a \emph{shared prior}, where the server provides the parameters for this prior, across clients. We believe that these views are complimentary to ours and combining the two is an interesting direction that we leave for future work.

Reducing communication costs is a well-known and explored topic in federated learning. \fedsparse{} has close connections to federated dropout~\citep{caldas2018expanding}, as the latter can be understood via a similar hierarchical model, where gates $\bv{z}$ are global and have a fixed probability $\bs{\theta}$ for both the prior and the approximate posterior. 
Compared to federated dropout, \fedsparse{} allow us to optimize the dropout rates to the data, such that they satisfy a given accuracy / sparsity trade-off, dictated by the hyperparameter $\lambda_0$. Another benefit of our EM perspective is that it clarifies that the server can perform gradient-based optimization. As a result, we can harvest the large literature on efficient distributed optimization~\citep{lin2017deep,bernstein2018signsgd,wangni2018gradient,yu2019double}, which involves gradient quantization, sparsification, and more general compression. On this front, there have also been other works that aim to reduce the communication cost in FL via such approaches~\citep{sattler2019robust,han2020adaptive}. In general, such approaches can be orthogonal to \fedsparse{} and exploring how they can be incorporated is a promising avenue for future research. Besides parameter / gradient compression, there are also works that reduce communication costs via other means.~\cite{karimireddy2020scaffold} reduces the number of training rounds necessary via a better optimization procedure and~\cite{li2020page} reduces the amount of gradient queries by reusing gradients from previous iterations. Both of these are orthogonal to \fedsparse{}, which decreases communication costs by reducing the number of parameters to transmit. Therefore, they could be combined for even more savings. As an example, we provide an experiment in Appendix~\ref{sec:scaffold_fedsparse} where we show the benefits of combining \fedsparse{} with~\cite{karimireddy2020scaffold}. 

%% file: sections/experiments.tex
\section{Experiments}
We verify in three tasks whether \fedsparse{} leads to similar or better global models compared to \fedavg{} while providing reductions in communication costs and efficient models. We include two baselines that also reduce communication costs by sparsifying the model; for the first, we consider the federated dropout procedure from~\cite{caldas2018expanding}, which we refer to as \feddrop{}, and for the second, we implement a variant of \fedavg{} where the clients locally perform group $L_1$ regularization \cite{wen2016learning} in order to sparsify their copy of the model.
The latter, which we refer to as \fedlone{}, is closer to what we do in \fedsparse{} and can similarly reduce the communication costs by employing a soft thresholding step on the model before communicating. For each task, we present the results for \fedsparse{} with regularization strengths that target three sparsity levels: low, mid, and high. For the \feddrop{} baseline, we experiment with multiple combinations of dropout probabilities for the convolutional and fully connected layers. For \fedlone{}, we considered a variety of regularization strengths. For each of these, we report the setting that performs best in terms of accuracy / communication trade-off. 

The first task we consider is a federated version of CIFAR10 classification where we partition the data among 100 shards in a non-i.i.d. way following~\cite{hsu2019measuring} and train a LeNet-5 convolutional architecture~\citep{lecun1998gradient} for 1k rounds. For the second task, we consider the 500 shard federated version of CIFAR100 classification from~\cite{reddi2020adaptive}, with a ResNet20 which we optimize for 6k rounds. For the final task, we considered the non-i.i.d. Femnist classification and we use the same configuration as CIFAR10, but we optimize the model for 6k rounds. More details can be found in Appendix~\ref{sec:exp_details}. 

We evaluate \fedsparse{} and the baselines on two metrics that highlight the tradeoffs between accuracy and communication costs. On both metrics, the x-axis represents the total communication cost incurred up until that point, and the y-axis represents two distinct model accuracies. The first one corresponds to the accuracy of the global model on the union of the shard test sets, whereas the second one corresponds to the average accuracy of the shard-specific ``local models'' on the shard-specific test sets. The ``local model'' on each shard is the model configuration that the shard last communicated to the server and serves as a proxy for the personalized model performance on each shard. The latter metric is motivated from the meta-learning~\citep{jiang2019improving} and hierarchical model view of federated learning and corresponds to using the local posteriors for prediction on the local test set instead of the server-side priors. All experiments were implemented using PyTorch \cite{NEURIPS2019_9015}.

\subsection{Experimental results}
The results from our experiments can be found in the following table and figures, where we report the average metrics and standard errors of each from 3 different random seeds. Overall, we observed that the \fedsparse{} models achieve their final sparsity ratios early in training, \emph{i.e.}, after 30-50 rounds, which quickly reduces the communication costs for each round (Appendix \ref{sec:results_sparsity}).

\begin{table*}[ht]
\centering
\begin{minipage}[b]{.5\textwidth}
\centering
\caption{Cifar10}
\addtolength{\tabcolsep}{-4pt}    
\resizebox{\textwidth}{!}{
\begin{tabular}{lcccc}
\toprule
Method & \makecell{Global \\ acc.} & \makecell{Local \\ acc.} & \makecell{GB \\ comm.} & Sparsity \\
\midrule
\fedavg{} & $69.7_{\pm0.3}$ & $86.2_{\pm0.2}$ & $65_{\pm0}$ & - \\
\feddrop{} & $69.9_{\pm0.7}$ & $84.9_{\pm0.2}$ & $34_{\pm0}$ & - \\
\fedlone{} & $62.1_{\pm0.8}$ & $82.5_{\pm0.3}$ & $36_{\pm1}$ & $61.8_{\pm1.8}$ \\
\midrule
\fedsparse{}, low & $68.9_{\pm0.5}$ & $86.9_{\pm0.4}$ & $52_{\pm0}$ & $21.4_{\pm0.4}$ \\
\fedsparse{}, mid & $68.9_{\pm0.4}$ & $87.6_{\pm0.1}$ & $35_{\pm0}$ & $46.8_{\pm0.4}$ \\
\fedsparse{}, high & $68.4_{\pm0.5}$ & $86.9_{\pm0.2}$ & $23_{\pm1}$ & $66.5_{\pm0.8}$ \\
\midrule
\bottomrule
\end{tabular}
}
\end{minipage}%
\begin{minipage}[b]{.5\textwidth}
\centering
\caption{Cifar100}
\addtolength{\tabcolsep}{-4pt}    
\resizebox{\textwidth}{!}{
\begin{tabular}{lcccc}
\toprule
Method & \makecell{Global \\ acc.} & \makecell{Local \\ acc.} & \makecell{GB \\ comm.} & Sparsity \\
\midrule
\fedavg{} & $40.6_{\pm1.0}$ & $61.5_{\pm0.6}$ & $124_{\pm0}$ & - \\
\feddrop{} & $36.1_{\pm0.5}$ & $57.8_{\pm0.3}$ & $112_{\pm0}$ & - \\
\fedlone{} & $23.6_{\pm0.7}$ & $45.8_{\pm0.3}$ & $118_{\pm1}$ & $8.0_{\pm1.0}$ \\
\midrule
\fedsparse{}, low & $40.5_{\pm0.7}$ & $61.8_{\pm0.4}$ & $121_{\pm0}$ & $1.6_{\pm0.1}$ \\
\fedsparse{}, mid & $31.7_{\pm1.0}$ & $54.8_{\pm0.6}$ & $58_{\pm1}$ & $53.8_{\pm0.6}$ \\
\fedsparse{}, high & $23.8_{\pm2.0}$ & $48.5_{\pm1.5}$ & $43_{\pm2}$ & $66.9_{\pm1.3}$ \\
\midrule
\bottomrule
\end{tabular}
}
\end{minipage}%
\label{tab:acc_comm_res}
\end{table*}

\begin{wraptable}{l}{7cm}
\centering
\small
\caption{Femnist}
\addtolength{\tabcolsep}{-4pt}    
\begin{tabular}{lcccc}
\toprule
Method & \makecell{Global \\ acc.} & \makecell{Local \\ acc.} & \makecell{GB \\ comm.} & Sparsity \\
\midrule
\fedavg{} & $85.8_{\pm0.2}$ & $90.8_{\pm0.0}$ & $272_{\pm0}$ & - \\
\feddrop{} & $84.7_{\pm0.2}$ & $87.6_{\pm0.0}$ & $168_{\pm0}$ & - \\
\fedlone{} & $82.0_{\pm0.3}$ & $85.9_{\pm0.0}$ & $187_{\pm2}$ & $45.5_{\pm1.2}$ \\
\midrule
\fedsparse{}, low & $85.4_{\pm0.2}$ & $90.8_{\pm0.0}$ & $270_{\pm0}$ & $1.0_{\pm0.2}$ \\
\fedsparse{}, mid & $85.1_{\pm0.3}$ & $90.6_{\pm0.0}$ & $148_{\pm2}$ & $45.6_{\pm0.7}$ \\
\fedsparse{}, high & $84.1_{\pm0.4}$ & $89.6_{\pm0.1}$ & $102_{\pm3}$ & $62.8_{\pm1.1}$ \\
\midrule
\bottomrule
\end{tabular}
\label{tab:acc_comm_res}
\end{wraptable}

\begin{figure*}[ht!]
\centering
 \begin{subfigure}[b]{0.32\linewidth}
  \centering
  \includegraphics[width=\linewidth]{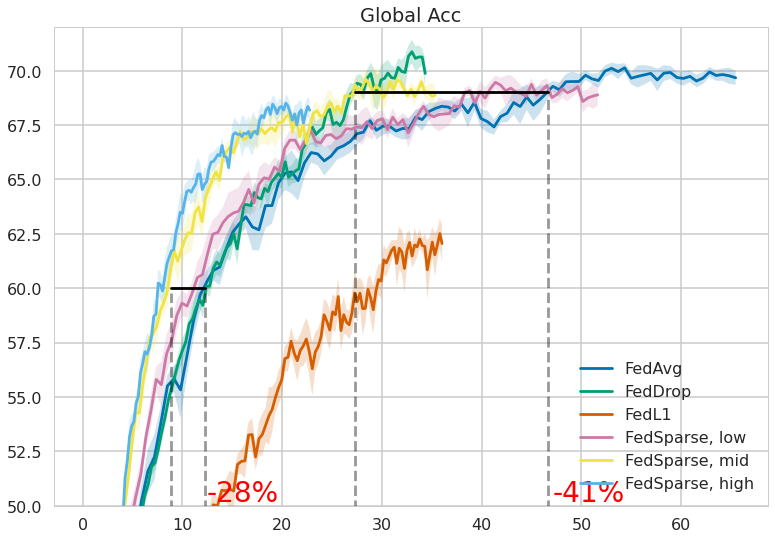}
  \caption{Cifar10}
 \end{subfigure}%
 ~
 \begin{subfigure}[b]{0.32\linewidth}
 \centering
  \includegraphics[width=\linewidth]{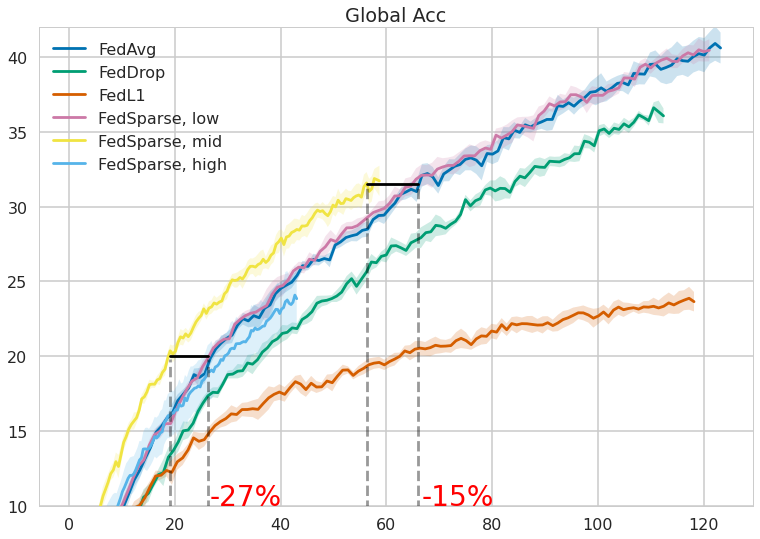}
  \caption{Cifar100}
  \label{fig:cifar100GB}
 \end{subfigure}%
 ~
  \begin{subfigure}[b]{0.32\linewidth}
 \centering
  \includegraphics[width=\linewidth]{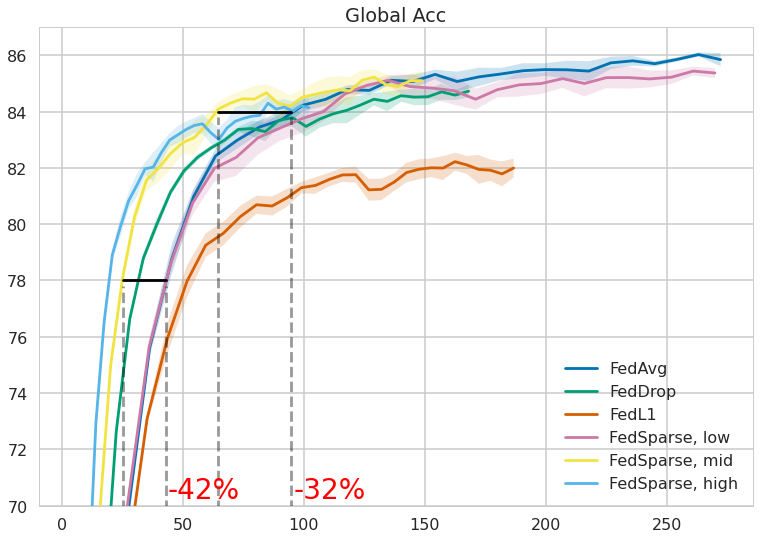}
  \caption{Femnist}
 \end{subfigure}
 \centering
 \begin{subfigure}[b]{0.32\linewidth}
  \centering
  \includegraphics[width=\linewidth]{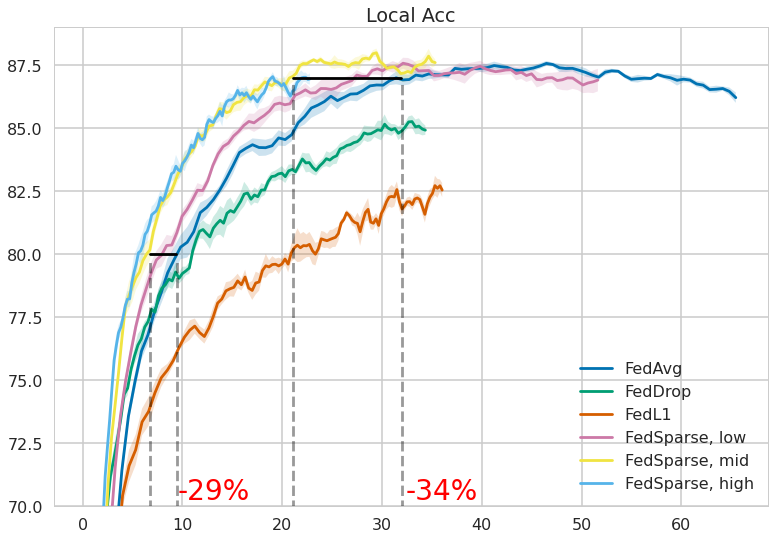}
  \caption{Cifar10}
  \label{fig:res_cifar10_steps}
 \end{subfigure}%
 ~
 \begin{subfigure}[b]{0.32\linewidth}
 \centering
  \includegraphics[width=\linewidth]{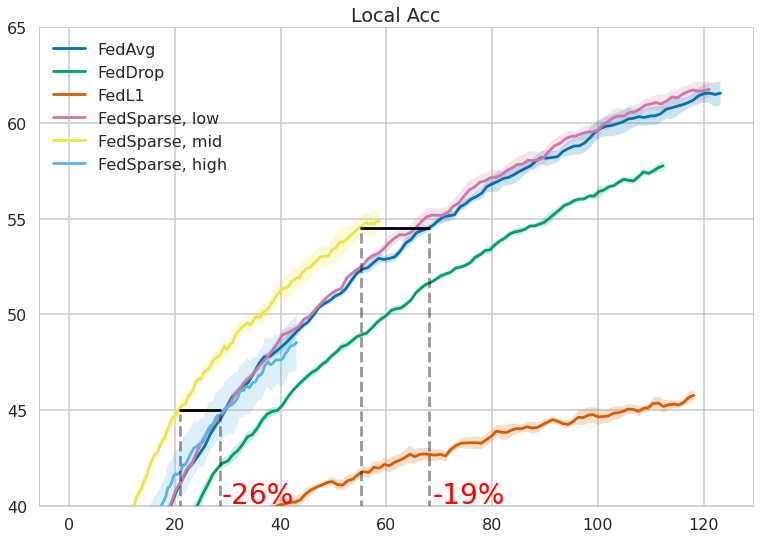}
  \caption{Cifar100}
 \end{subfigure}%
 ~
  \begin{subfigure}[b]{0.32\linewidth}
 \centering
  \includegraphics[width=\linewidth]{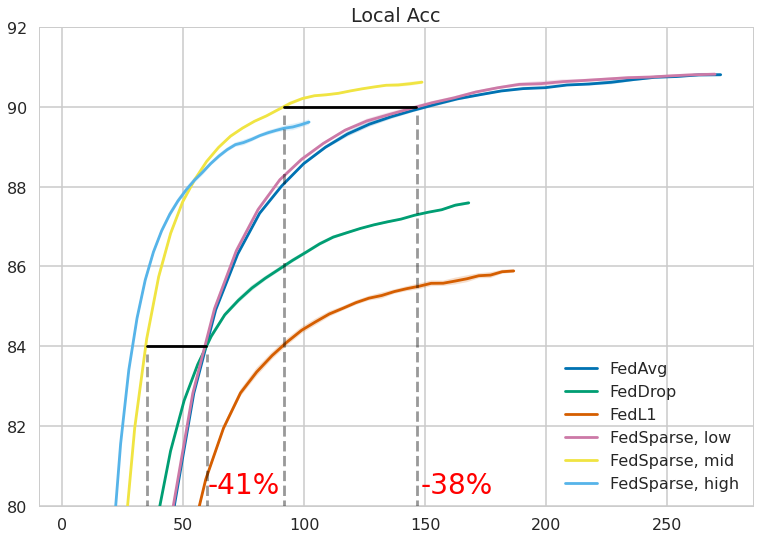}
  \caption{Femnist}
  \label{fig:femnist_steps}
 \end{subfigure}
 \caption{Global model accuracy (top row) and average local model accuracy (bottom row) across all clients (y-axis) as a function of the amount of GB communicated (top row). Best viewed in color.} 
 \label{fig:res_acc_vs_comm}
\end{figure*}

We can see that for CIFAR 10, the \fedsparse{} models with medium ($\sim$47\%) and high ($\sim$66\%) sparsity outperform all other methods for small communications budgets on the global accuracy front but are eventually surpassed by \feddrop{} on higher budgets. However, on the local accuracy front, we see that the \fedsparse{} models perform better than both baselines, achieving, \emph{e.g.}, $\sim$ 87\% local accuracy with $\sim$34\% less communication compared to \fedavg{}. Overall, judging the final performance only, we see that \feddrop{} reaches the best accuracy on the global model, but \fedsparse{} reaches the best accuracy in the local models. \fedlone{} was able to learn a sparse model that was comparable in size to the \fedsparse{} `high' setting, however, the final accuracies, both the global and the local, were worse than all of the other methods.

On CIFAR 100, the differences are less pronounced, as the models did not fully converge for the maximum number of rounds we use. Nevertheless, we still observe similar patterns; for small communication budgets, the sparser models are better for both the global and local accuracy as, \emph{e.g.}, they can reach $\sim$32\% global accuracy while requiring $\sim$15\% less communication than \fedavg{}. 

Finally, for the Femnist task, which is also the more communication-intensive due to having $\sim$3.5k shards, we see that the \fedsparse{} algorithm improves upon both \feddrop{} and \fedavg{}. More specifically, in the medium sparsification setting, it can reach $\sim$85\% global accuracy and $\sim$90\% local accuracy while requiring $\sim$32\% and $\sim$38\% less communication compared to \fedavg{} respectively. Judging by the final accuracy, both \fedavg{} and \fedsparse{} with the low setting reached similar global and local model performance. This is expected, given that that particular \fedsparse{} setting leads to only 1\% sparsity. \fedlone{} did not outperform any of the other methods in terms of the final accuracies.

%% file: sections/conclusion.tex
\section{Conclusion}
In this work, we adopted an interpretation of server-orchestrated federated learning as a hierarchical model where the server provides the parameters of a prior distribution over the parameters of client-specific models. We then showed how the \fedavg{} algorithm, the standard in federated learning, corresponds to applying the hard-EM algorithm to a hierarchical model that uses Gaussian priors. Through this perspective, we bridged several recent works on federated learning as well as connect \fedavg{} to meta-learning.  
As a straightforward extension stemming from this view, we proposed a hierarchical model with sparsity-inducing priors. By applying hard-EM on this new model, we obtained \fedsparse{}, a generalization of \fedavg{} that can learn sparse neural networks in the federated learning setting. Empirically, we showed that \fedsparse{} can learn sparse neural networks, which, besides being more efficient, can also significantly reduce the communication costs without decreasing performance - both of which are of great practical importance in ``cross-device'' federated learning. Of equal practical importance is hyperparameter resilience. In training models with \fedsparse{} we observed a high sensitivity to hyperparameters related to initial sparsity thresholds and the necessity to downscale cross-entropy terms as discussed in Appendix \ref{sec:exp_details}. A successful application of \fedsparse{} therefore relies on careful tuning and provides ample opportunity for improvement in future work.

%% file: sections/appendix.tex
\appendix

\section{Experimental details}
\label{sec:exp_details}
 For all of the three tasks we randomly select 10 clients without replacement in a given round but with replacement across rounds. For the local optimizer of the weights we use stochastic gradient descent with a learning rate of 0.05, whereas for the global optimizer we use Adam~\citepappendix{kingma2014adam} with the default hyperparameters provided in~\citepappendix{kingma2014adam}. For the pruning thresholds in \fedsparse{} we used the Adamax~\citepappendix{kingma2014adam} optimizer with $1e-3$ learning rate at the shard level and the Adamax optimizer with $1e-2$ learning rate at the server. For all three of the tasks we used $E=1$ with a batch size of 64 for CIFAR10 and 20 for CIFAR100 and Femnist. It should be noted that for all the methods we performed gradient based optimization using the difference gradient for the weights~\citepappendix{reddi2020adaptive} instead of averaging.
 \nocite{kingma2014adam}

For the \feddrop{} baseline, we used a very small dropout rate of 0.01 for the input and output layer and tuned the dropout rates for convolutional and fully connected layers separately in order to optimize the accuracy / communication tradeoff. For convolutional layers we considered rates in $\{0.1, 0.2, 0.3\}$ whereas for the fully connected layers we considered rates in $\{0.1, 0.2, 0.3, 0.4, 0.5\}$. For CIFAR10 we did not employ the additional dropout noise at the shard level, since we found that it was detrimental for the \feddrop{} performance. Furthermore, for Resnet20 on CIFAR100 we did not apply federated dropout at the output layer. For CIFAR10 the best performing dropout rates were 0.1 for the convolutional and 0.5 for the fully connected, whereas for CIFAR100 it was 0.1 for the convolutional. For Femnist, we saw that a rate of 0.2 for the convolutional and a rate of 0.4 for the fully connected performed better.

For \fedsparse{}, we initialized $\bv{v}$ such that the thresholds $\bs{\tau}$ lead to $\bs{\theta} = 0.99$ initially, \emph{i.e.} we started from a dense model. The temperature for the sigmoid in the parameterization of the probabilities was set to $T = 0.001$. Furthermore, we downscaled the cross-entropy term between the client side probabilities, $\bs{\pi}_s$, and the server side probabilities, $\bs{\theta}$ by mutltiplying it with $1e-4$. Since at the beginning of each round we were always initializing $\bs{\pi}_S = \bs{\theta}$ and we were only optimizing for a small number of steps before synchronizing, we found that the full strength of the cross-entropy was not necessary. Furthermore, for similar reasons, \emph{i.e.} we set $\bv{w}_s = \bv{w}$ at the beginning of each round, we also used $\lambda = 0$ for the drift term $\frac{\lambda}{2} \bs{\pi}_{sj} (\bv{w}_s - \bv{w})^2$. The remaining hyperparameter $\lambda_0$ dictates how sparse the final model will be. For the LeNet-5 model the $\lambda_0$'s we report are $\{5e-7, 5e-6, 5e-5\}$ for the ``low'', ``mid'' and ``high'' settings respectively, which were optimized for CIFAR10 and used as-is for Femnist. For CIFAR100 and Resnet20, we did not perform any pruning for the output layer and the $\lambda_0$'s for the ``low'', ``mid'' and ``high'' settings were $\{5e-7, 5e-6, 5e-5\}$ respectively. These were chosen so that we obtain models with comparable sparsity ratios as the one on CIFAR10.

The \fedlone{} baseline was implemented by using a group $L_1$ regularizer~\citeappendix{wen2016learning} at the client level, \emph{i.e.}, the local loss was:
\begin{align}
    \mathcal{L}_s(\mathcal{D}_s, \bv{w}) := \frac{1}{N_s}\sum_{i=1}^{N_s} L(\mathcal{D}_{si}, \bv{w}) + \lambda \sum_g \sqrt{\|\bv{w}_g\|_2^2},
\end{align}
where $\bv{w}_g$ was the vector of parameters associated with group $g$. To have similar sparsity patterns to \fedsparse{}, each of the groups of parameters were chosen to be the weights associated with a specific neuron (in case of a fully connected layer) or a specific feature map (in case of a convolutional layer). In order to realize the sparse communication, we employed a thresholding operator before each client and the server communicated their parameters:
\begin{align}
    \hat{\bv{w}}_g = \bv{w}_g z_g, \qquad z_g = 
    \begin{cases}
    0, \enskip \text{if } \sqrt{\|\bv{w}_g\|_2^2} \leq \lambda \\
    1, \enskip \text{otherwise}
    \end{cases}
\end{align}
We empirically found that this operator resulted into better overall model performance, compared to the more traditional soft-thresholding one, which also ``shrinks'' the non-pruned parameters. The server used the traditional parameter difference between the communicated, sparse, local model and the current server model for gradient based optimization. To obtain the results in the main text, we used $\lambda=2e-3$ for CIFAR10 and CIFAR100, whereas for Femnist we used $\lambda=1e-3$. These values were chosen so that we obtain comparable sparsities to the ones from \fedsparse{}.

Most of the experiments were run on an Nvidia RTX 2080Ti GPU, and the hyperparameter optimization was performed with several Nvidia V100 GPU's on an internal cluster over the span of a month.

\section{Evolution of sparsity}
\label{sec:results_sparsity}
We show the evolution of the sparsity ratios for all tasks and configurations in the following plot. We can see that in all settings the model attains its final sparsity quite early in training (i.e., before $100$ rounds) in the case of \fedsparse{} whereas with \fedlone{} the sparsity is achieved much later in training. As a result, the communication savings (of the overall training procedure) with \fedlone{} are not as much as the ones from \fedsparse{}.

\begin{figure}[ht!]
\begin{subfigure}[b]{0.32\linewidth}
 \centering
  \includegraphics[width=\linewidth]{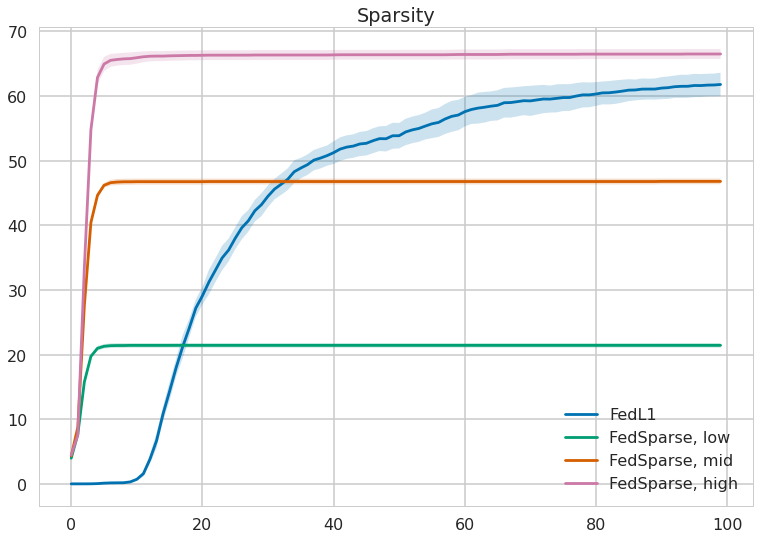}
  \caption{CIFAR10 sparsity}
 \end{subfigure}
 ~
\begin{subfigure}[b]{0.32\linewidth}
 \centering
  \includegraphics[width=\linewidth]{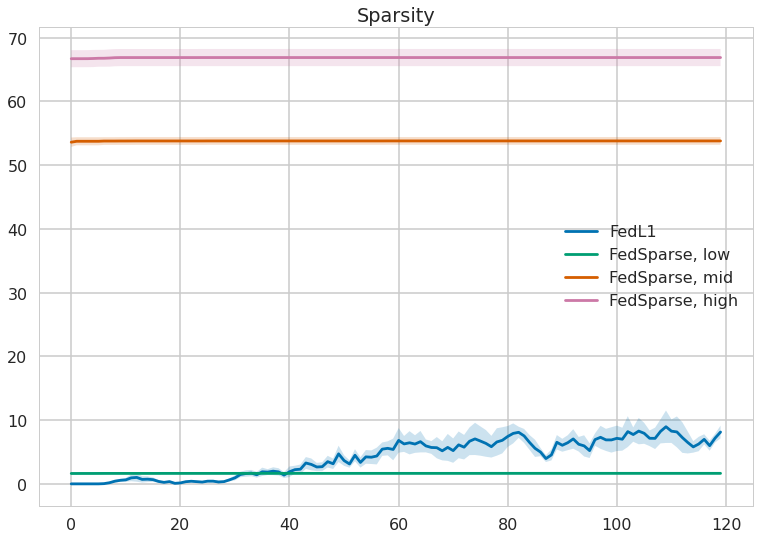}
  \caption{CIFAR100 sparsity}
 \end{subfigure}
   ~
 \begin{subfigure}[b]{0.32\linewidth}
 \centering
  \includegraphics[width=\linewidth]{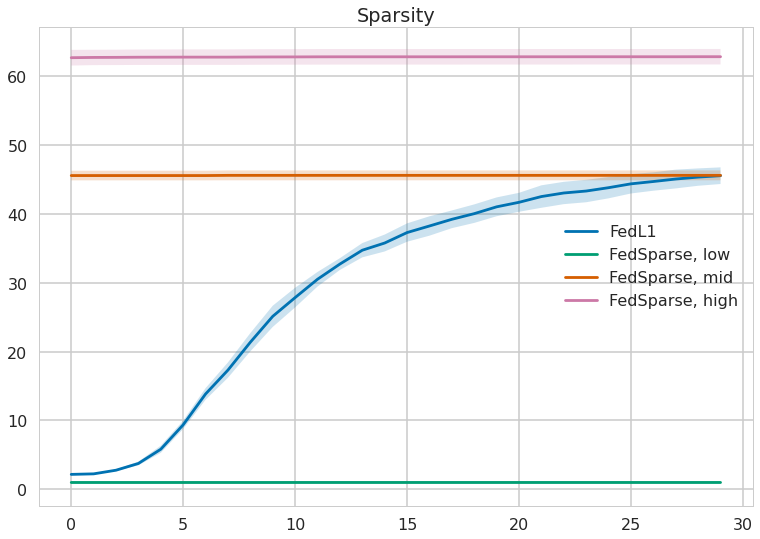}
  \caption{Femnist sparsity}
 \end{subfigure}
 \end{figure}

\section{Sparsity alignment between clients and server}
\begin{figure*}[h!]
\centering
 \begin{subfigure}[b]{0.32\linewidth}
  \centering
  \includegraphics[width=\linewidth]{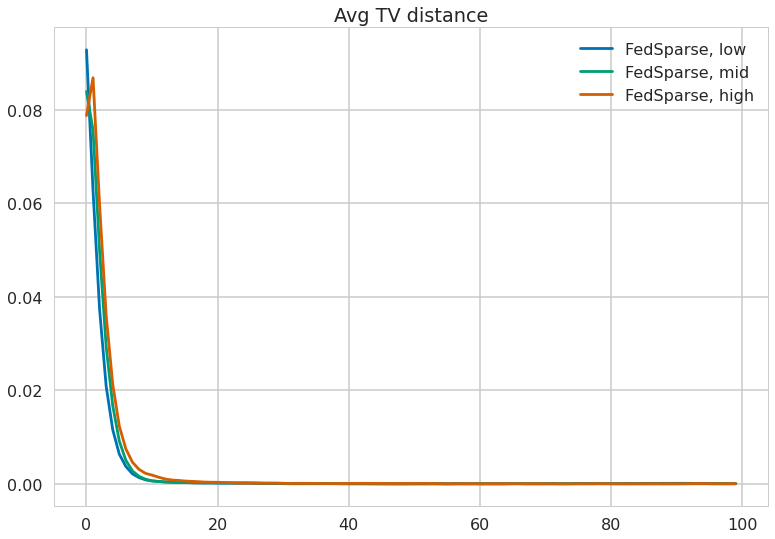}
  \caption{Cifar10}
 \end{subfigure}%
 ~
 \begin{subfigure}[b]{0.32\linewidth}
 \centering
  \includegraphics[width=\linewidth]{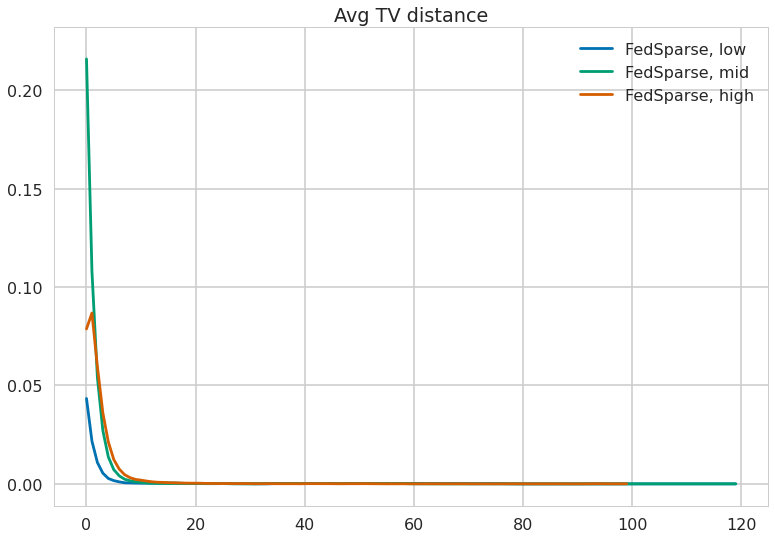}
  \caption{Cifar100}
 \end{subfigure}%
 ~
  \begin{subfigure}[b]{0.32\linewidth}
 \centering
  \includegraphics[width=\linewidth]{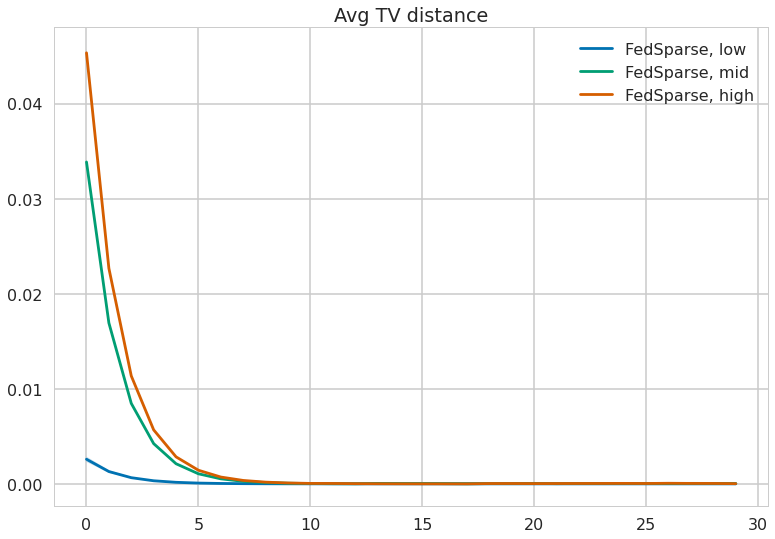}
  \caption{Femnist}
 \end{subfigure}
 
 \centering
 \begin{subfigure}[b]{0.32\linewidth}
  \centering
  \includegraphics[width=\linewidth]{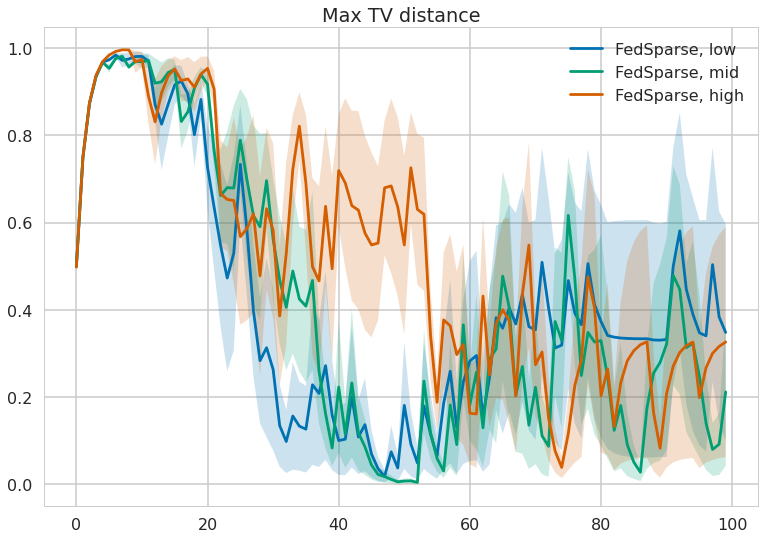}
  \caption{Cifar10}
 \end{subfigure}%
 ~
 \begin{subfigure}[b]{0.32\linewidth}
 \centering
  \includegraphics[width=\linewidth]{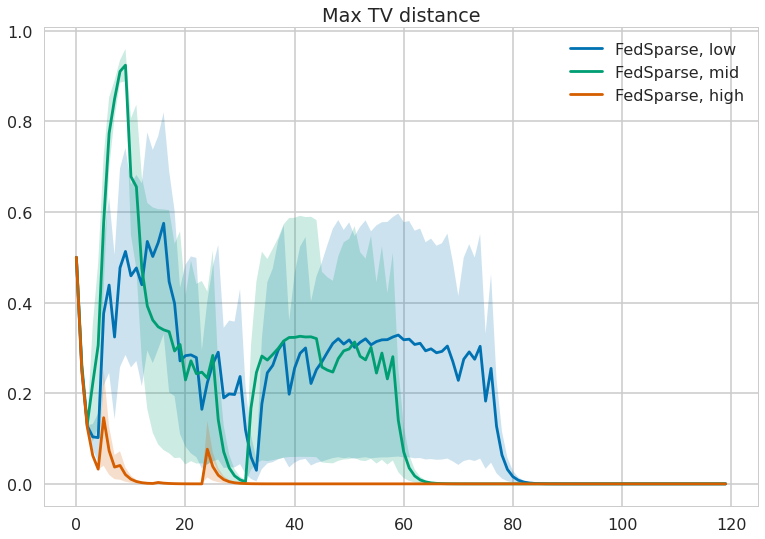}
  \caption{Cifar100}
 \end{subfigure}%
 ~
  \begin{subfigure}[b]{0.32\linewidth}
 \centering
  \includegraphics[width=\linewidth]{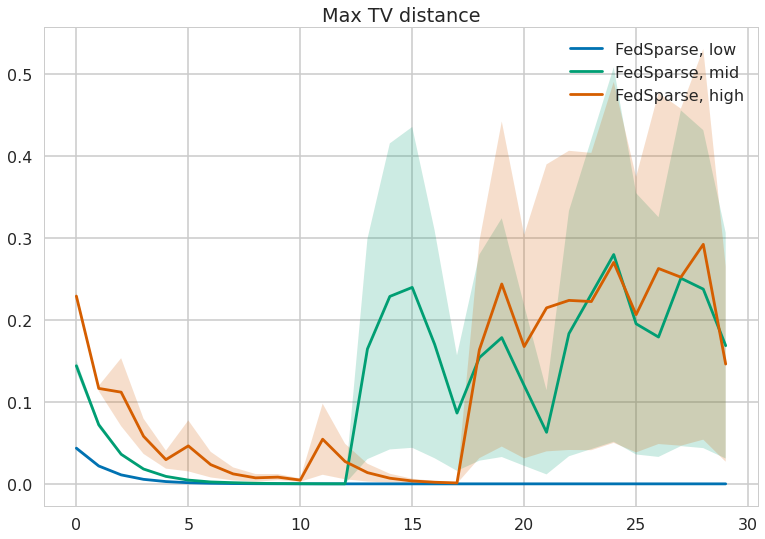}
  \caption{Femnist}
 \end{subfigure}
 \caption{Average (top row) and maximum (bottom row) total variation distance for all tasks and \fedsparse{} settings.} 
 \label{fig:tv_res}
\end{figure*}

A reasonable question to ask is whether the local sparsity patterns between clients ``align''; if the clients do not agree on which parameters to prune, the server model will not be sparse and therefore we won't get as much communication cost benefit. To test where such an issue happens in practice with \fedsparse{} we implemented two metrics that track the average and maximum total variation distance between the gating distributions (\emph{i.e.}, those that control the sparsity patterns) of the clients and the server. The total variation distance corresponds to the absolute difference between the client and server probabilities for the binary gates, \emph{i.e.},
\begin{align}
    \text{TV}\left(q_{\pi_s}(z_s), p_\theta(z_s)\right) = \left|q_{\pi_s}(z_s = 1) - p_\theta(z_s=1)\right| = \left|\pi_s - \theta\right|
\end{align}
The resulting plots over the course of training for all tasks and \fedsparse{} settings can be found in Fig.~\ref{fig:tv_res}.
We can clearly see that the clients and the server quickly agree on the general sparsity pattern as the average total variation distance becomes almost zero on all settings. We can attribute this to the ``resetting'' of the local $q_{\pi_s}(z_s)$ to $p_\theta(z_s)$ configuration that happens at the beginning of each round along with the extra cross-entropy loss between $q_{\pi_s}(z_s), p_\theta(z_s)$ that \fedsparse{} has. We can also see that the maximum total variation distance takes much longer to drop, meaning that there are a few parameter groups where the probability of keeping them at the clients' sides is different than the one at the server. We attribute this to the model "specializing" the sparsity pattern to the characteristics of the client dataset. We can also see however, that the maximum total variation distance does drop as training progresses, showing that the sparsity pattern specialization does decrease (although it never becomes zero). 

\section{Additional results}
\paragraph{Convergence plots in terms of communication rounds.}
In order to understand whether the extra noise is detrimental to the convergence speed of \fedsparse{}, we plot the validation accuracy in terms of communication rounds for all tasks and baselines. As it can be seen, there is no inherent difference before \fedsparse{} starts pruning. This happens quite early in training for CIFAR 100 thus it is there where we observe the most differences. \fedlone{} is overall worse than the other methods as the inclusion of the $L_1$ penalty hurts model performance.

\begin{figure}[ht!]
\begin{subfigure}[b]{0.32\linewidth}
 \centering
  \includegraphics[width=\linewidth]{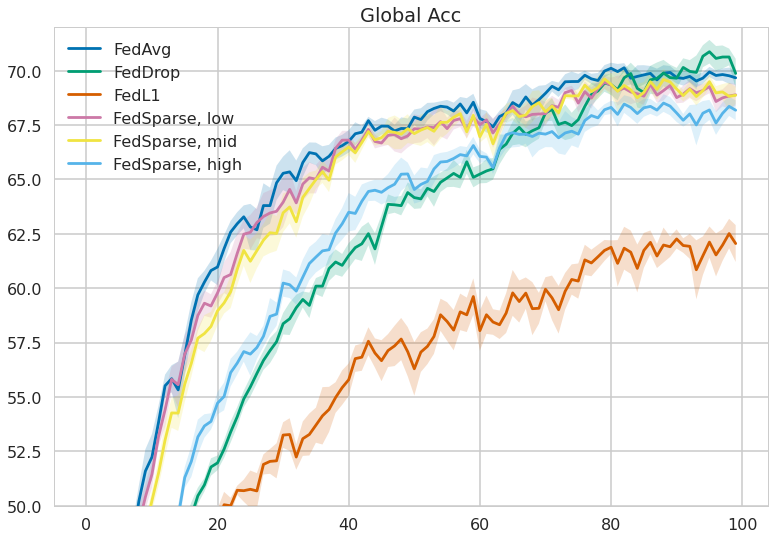}
  \caption{CIFAR10}
 \end{subfigure}
 ~
\begin{subfigure}[b]{0.32\linewidth}
 \centering
  \includegraphics[width=\linewidth]{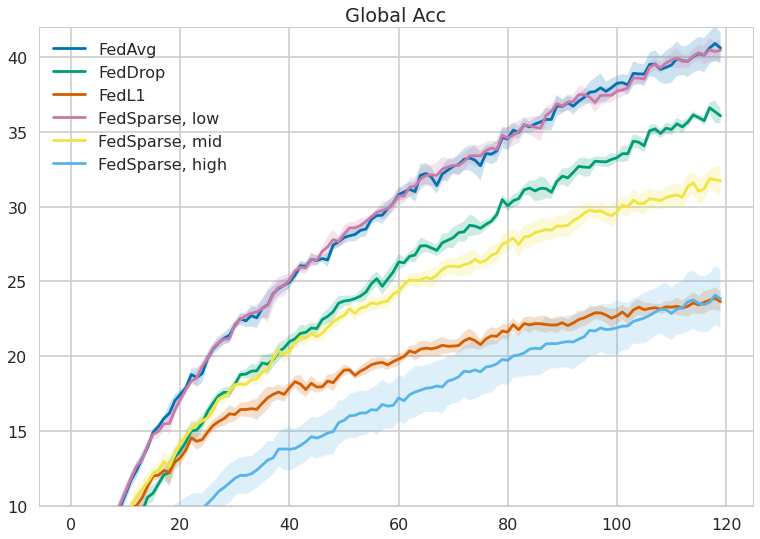}
  \caption{CIFAR100}
 \end{subfigure}
   ~
 \begin{subfigure}[b]{0.32\linewidth}
 \centering
  \includegraphics[width=\linewidth]{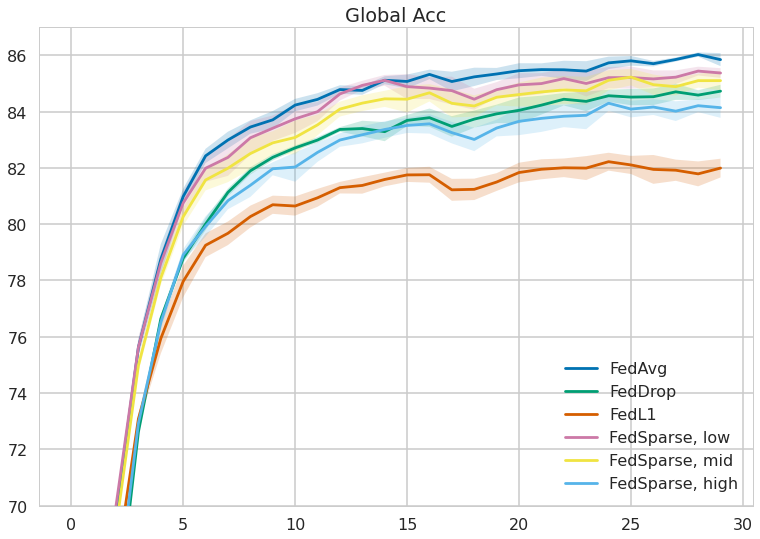}
  \caption{Femnist}
 \end{subfigure}
 \caption{Evolution of the validation accuracy in terms of communication rounds.}
 \end{figure}

\paragraph{Impact of server side pruning.} In order to understand whether server side pruning is harmful for convergence, we plot both the global and average local validation accuracy on CIFAR 10 for the ``mid'' setting of \fedsparse{} with and without server side pruning enabled. As we can see, there are no noticeable differences and in fact, pruning results into a slightly better overall performance.
\begin{figure}[ht!]
\begin{subfigure}[b]{0.49\linewidth}
 \centering
  \includegraphics[width=\linewidth]{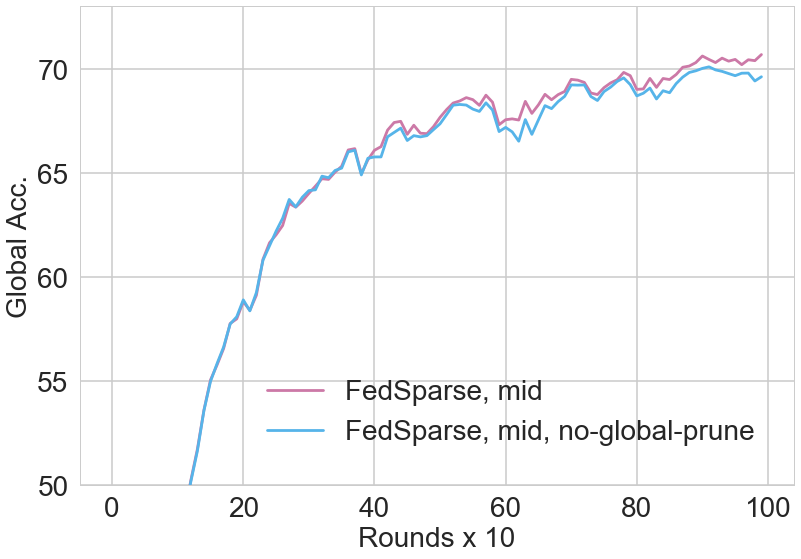}
  \caption{CIFAR10 global val. acc.}
 \end{subfigure}
 ~
\begin{subfigure}[b]{0.49\linewidth}
 \centering
  \includegraphics[width=\linewidth]{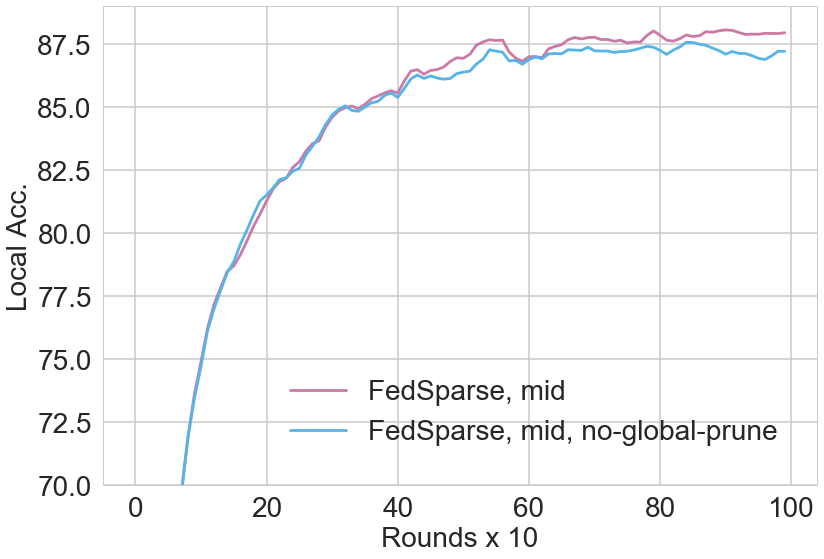}
  \caption{CIFAR100 local val. acc.}
 \end{subfigure}
 \caption{Evolution of the validation accuracy in terms of communication rounds with and without server side pruning.}
 \end{figure}

\section{Combining SCAFFOLD with \fedsparse{}}
\label{sec:scaffold_fedsparse}
\begin{figure}[ht!]
\begin{subfigure}[b]{0.49\linewidth}
 \centering
  \includegraphics[width=\linewidth]{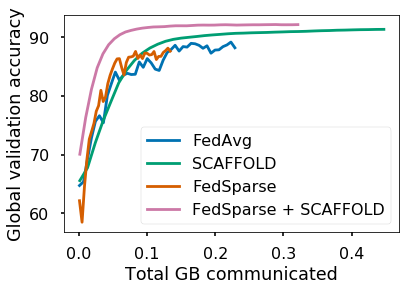}
  \caption{MNIST global val. acc.}
 \end{subfigure}
 ~
\begin{subfigure}[b]{0.49\linewidth}
 \centering
  \includegraphics[width=\linewidth]{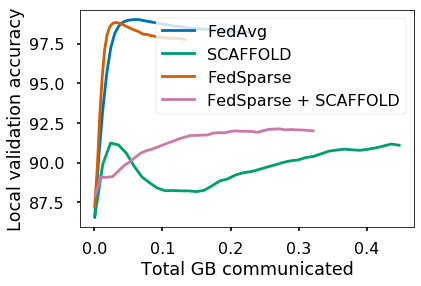}
  \caption{MNIST local val. acc.}
 \end{subfigure}
 \caption{Evolution of the validation accuracy in terms of the GB communicated for \fedavg{}, \scaffold{}, \fedsparse{} and \fedsparse{} with \scaffold{}.}\label{fig:scaffold_fedsparse_res}
 \end{figure}

In order to show that techniques such as~\citeappendix{karimireddy2020scaffold,li2020page} are orthogonal to \fedsparse{}, we implement \scaffold{}~\citeappendix{karimireddy2020scaffold} on a simple convex problem (following their theory); logistic regression on a (extreme) non-iiid split of the MNIST dataset into 100 users (each user had data from a single class) trained for 400 rounds. For the combination of \fedsparse{} and \scaffold{}, we used the \scaffold{} procedure for the weights. The results can be seen in Figure~\ref{fig:scaffold_fedsparse_res}.

When measuring the global model accuracy, we observe the benefits of \scaffold{} compared to \fedavg{}. Nevertheless, the “local model“ accuracy (as defined in the main text) was worse, probably due to the control variates leading to a model that is less fine-tuned on the local data-sets. 
By comparing these results to the ones with \fedsparse{}, we can see that indeed \fedsparse{} is orthogonal to \scaffold{}, so one can combine both methods to get the best of both worlds. The model sparsity for \fedsparse{} was around 35\% when including \scaffold{} and 47\% without. It should be noted that \scaffold{} doubles the communication cost per round (compared to \fedavg{}) due to transmitting the control variates.

On the neural network architectures of the main text, the picture was not as clear since \scaffold{} lead to unstable optimization, which only stabilised after we ''dampened” the contribution from the control variates to the local training procedure (thus not providing benefits). 
Since in \citeappendix{karimireddy2020scaffold} it is mentioned that on non-convex problems ''much more extensive experiments (beyond current scope) are needed before drawing conclusions”, we did not pursue this further.

\section{Correspondence between single step EM and gradient ascent}
\label{app:singleStepEMGradAscent}
With the addition of the auxiliary variables $\bs{\phi}_s$ we have that the overall objective for the server becomes
\begin{align}
    \argmax_\bv{w} \frac{1}{S}\sum_{s=1}^S \log \int p(\mathcal{D}_s | \bs{\phi}_s) p(\bs{\phi}_s | \bv{w})\mathrm{d}\bs{\phi}_s. \label{eq:latent_fedavg}
\end{align}
By performing EM with a single gradient step for $\bv{w}$ in the M-step (instead of full maximization), we are essentially doing gradient ascent on the original objective at~\ref{eq:latent_fedavg}. To see this, we can take the gradient of Eq.~\ref{eq:latent_fedavg} w.r.t. $\bv{w}$ where $Z_s = \int p(\mathcal{D}_s |\bs{\phi}_s)p(\bs{\phi}_s | \bv{w})\mathrm{d}\bs{\phi}_s$
\begin{align}
    & \frac{1}{S}\sum_s \frac{1}{Z_s}\int p(\mathcal{D}_s | \bs{\phi}_s)\frac{\partial p(\bs{\phi}_s | \bv{w})}{\partial \bv{w}}\mathrm{d}\bs{\phi}_s =\\
    & \frac{1}{S}\sum_s \int \frac{p(\mathcal{D}_s | \bs{\phi}_s)p(\bs{\phi}_s | \bv{w})}{Z_s}\frac{\partial \log p(\bs{\phi}_s | \bv{w})}{\partial \bv{w}}\mathrm{d}\bs{\phi}_s =\\
    & \frac{1}{S}\sum_s \int p(\bs{\phi}_s | \mathcal{D}_s, \bv{w}) \frac{\partial \log p(\bs{\phi}_s | \bv{w})}{\partial \bv{w}}\mathrm{d}\bs{\phi}_s \label{eq:grad_em}
\end{align}
where to compute Eq.~\ref{eq:grad_em} we see that we first have to obtain the posterior distribution of the local variables $\bs{\phi}_s$ and then estimate the gradient for $\bv{w}$ by marginalizing over this posterior.

\section{Derivation of the local loss for \fedsparse{}}
\label{sec:local_loss}
Let $p(\bs{\phi}_{si} | \bv{w}_i, \bv{z}_{si} = 1) = \mathcal{N}(\bv{w}_i, 1/\lambda), p(\bs{\phi}_{si} | \bv{w}_i, \bv{z}_{si} = 0) = \mathcal{N}(0, 1/\lambda_2)$ and $q(\bs{\phi}_{si} | \bv{z}_{si} = 1) = \mathcal{N}(\bv{w}_{si}, \epsilon^2), q(\bs{\phi}_{si} | \bv{z}_{si} = 0) = \mathcal{N}(0, \epsilon^2)$. Furthermore, let $q(\bv{z}_{si}) = \text{Bern}(\bs{\pi}_{si})$. The local objective that stems from~\ref{eq:init_elbo_kl_w} can be rewritten as:
\begin{align}
    \argmax_{\bv{w}_{1:S}, \bs{\pi}_{1:S}}  \mathbb{E}_{q_{\bs{\pi}_s}(\bv{z}_s)q_{\bv{w}_s}(\bs{\phi}_s|\bv{z}_s)}\big[&\log p(\mathcal{D}_s| \bs{\phi}_s)\big] - \mathbb{E}_{q_{\bs{\pi}_s}(\bv{z}_s)}\big[KL(q_{\bv{w}_s}(\bs{\phi}_s| \bv{z}_s)||p(\bs{\phi}_s | \mathbf{w}, \bv{z}_s))\big] \nonumber\\ & \quad + \mathbb{E}_{q_{\bs{\pi}_s}(\bv{z}_s)}[\log p(\bv{z}_s | \bs{\theta})],\label{eq:elbo_kl_w}
\end{align}
where we omitted from the objective the entropy of the distribution over the local gates.

One of the quantities that we are after is
\begin{align}
    \mathbb{E}_{q(\bv{z}_{si})}[&KL(q(\bs{\phi}_{si} | \bv{z}_{si}) ||  p(\bs{\phi}_{si} | \bv{z}_{si}))] = \nonumber \\ &\bs{\pi}_{si} KL(\mathcal{N}(\bv{w}_{si}, \epsilon^2) || \mathcal{N}(\bv{w}_i, 1/\lambda)) + (1 - \bs{\pi}_{si})KL(\mathcal{N}(0, \epsilon^2) || \mathcal{N}(0, 1/\lambda_2)).
\end{align}
The KL term for when $\bv{z}_{si} = 1$ can be written as
\begin{align}
    KL(\mathcal{N}(\bv{w}_{si}, \epsilon^2) || \mathcal{N}(\bv{w}_i, 1/\lambda)) = - \frac{1}{2}\log\lambda - \log\epsilon + \frac{\lambda \epsilon^2}{2} - \frac{1}{2} + \frac{\lambda}{2}(\bv{w}_{si} - \bv{w}_i)^2.
\end{align}
The KL term for when $\bv{z}_{si} = 0$ can be written as
\begin{align}
    KL(\mathcal{N}(0, \epsilon^2) || \mathcal{N}(0, 1/\lambda_2)) = - \frac{1}{2}\log\lambda_2 - \log\epsilon + \frac{\lambda_2 \epsilon^2}{2} - \frac{1}{2}.
\end{align}
Taking everything together we thus have
\begin{align}
    \mathbb{E}_{q(\bv{z}_{si})}[KL(q(\bs{\phi}_{si} | \bv{z}_{si}) || p(\bs{\phi}_{si} | \bv{z}_{si}))] & = \frac{\lambda \bs{\pi}_{si}}{2}(\bv{w}_{si} - \bv{w}_i)^2  + \bs{\pi}_{si} (- \frac{1}{2}\log\lambda - \log\epsilon + \frac{\lambda \epsilon^2}{2} - \frac{1}{2}) + \nonumber \\ & \qquad (1 - \bs{\pi}_{si})(- \frac{1}{2}\log\lambda_2 - \log\epsilon + \frac{\lambda_2 \epsilon^2}{2} - \frac{1}{2})\\
    & = \frac{\lambda \bs{\pi}_{si}}{2}(\bv{w}_{si} - \bv{w}_i)^2 + \bs{\pi}_{si}\left(\frac{1}{2}\log\frac{\lambda_2}{\lambda} + \frac{\epsilon^2}{2}(\lambda - \lambda_2)\right) + C\\
    & \approx \frac{\lambda \bs{\pi}_{si}}{2}(\bv{w}_{si} - \bv{w}_i)^2 + \lambda_0\bs{\pi}_{si} + C,
\end{align}
where $\lambda_0 = \frac{1}{2}\log\frac{\lambda_2}{\lambda}$ and $\frac{\epsilon^2}{2}(\lambda - \lambda_2)$ was omitted due to $\epsilon^2 \approx 0$. In the appendix of~\citeappendix{louizos2017learning}, the authors argue about a hypothetical prior that results into needing $\lambda$ nats to transform that prior to the approximate posterior. Here we make this claim more precise and show that this prior is approximately equivalent to a mixture of Gaussians prior where the precision of the non-zero prior component $\lambda \rightarrow \epsilon$ (in order to avoid the $L_2$ regularization term) and the precision of the zeroth component $\lambda_2$ is equivalent to $\lambda \exp(2 \lambda_0)$, where $\lambda_0$ is the desired $L_0$ regularization strength. 

Furthermore, the cross-entropy from $q_{\bs{\pi}_s}(\bv{z}_s)$ to $p(\bv{z}_s | \bs{\theta})$ is straightforward to compute as
\begin{align}
    \mathbb{E}_{q_{\bs{\pi}_s}(\bv{z}_s)}[\log p(\bv{z}_s | \bs{\theta})] = \sum_j \left( \bs{\pi}_{sj} \log\bs{\theta}_j + (1 - \bs{\pi}_{sj})\log(1 - \bs{\theta}_j)\right).
\end{align}
By putting everything together we have that the local objective becomes
\begin{align}
    \argmax_{\bv{w}_s, \bs{\pi}_s} 
    \mathbb{E}_{q_{\bs{\pi}_s}(\bv{z}_s)}\left[\sum_i^{N_s}L(\mathcal{D}_{si}, \bv{w}_s \odot \bv{z}_s)\right] &- \frac{\lambda}{2} \sum_j \bs{\pi}_{sj}(\bv{w}_{sj} - \bv{w}_j)^2  - \lambda_0 \sum_j \bs{\pi}_{sj} \nonumber \\ & + \sum_j \left(\bs{\pi}_{sj}\log\bs{\theta}_j + (1 - \bs{\pi}_{sj}) \log(1 - \bs{\theta}_j)\right) + C.
\end{align}

\section{Local optimization of the binary gates}
\label{app:localgates}
We propose to rewrite the local loss in Eq.~\ref{eq:init_elbo_kl_w} to
\begin{align}
   \mathcal{L}_s(\mathcal{D}_s, \bv{w}, \bs{\theta}, \bs{\phi}_s, & \bs{\pi}_s) := \mathbb{E}_{q_{\bs{\pi}_s}(\bv{z}_s)}\bigg[\sum_i^{N_s}L(\mathcal{D}_{si}, \bv{w}_s \odot \bv{z}_s) - \lambda \sum_j \mathbb{I}[\bv{z}_{sj} \neq 0 ](\bv{w}_{sj} - \bv{w})^2 \nonumber \\ &\qquad - \lambda_0 \sum_j \mathbb{I}[\bv{z}_{sj} \neq 0] + \sum_j \left(\mathbb{I}[\bv{z}_{sj} \neq 0]\log\frac{\bs{\theta}_j}{1 - \bs{\theta}_j} + \log (1 - \bs{\theta}_j)\right)\bigg], \label{eq:local_lb_ss_alt}
\end{align}
and then replace the Bernoulli distribution $q_{\bs{\pi}_s}(\bv{z}_s)$ with a continuous relaxation, the hard-concrete distribution~\citepappendix{louizos2017learning}. Let the continuous relaxation be $r_{\bv{u}_s}(\bv{z}_s)$, where $\bv{u}_s$ are the parameters of the surrogate distribution. In this case the local objective becomes
\begin{align}
   \mathcal{L}_s(\mathcal{D}_s, \bv{w}, \bs{\theta}, & \bs{\phi}_s, \bv{u}_s) := \mathbb{E}_{r_{\bv{u}_s}(\bv{z}_s)}\bigg[\sum_i^{N_s}L(\mathcal{D}_{si}, \bv{w}_s \odot \bv{z}_s)\bigg] - \lambda \sum_j R_{\bv{u}_{sj}}(\bv{z}_{sj} > 0)(\bv{w}_{sj} - \bv{w})^2 \nonumber \\ & - \lambda_0 \sum_j R_{\bv{u}_{sj}}(\bv{z}_{sj} > 0) + \sum_j\left( R_{\bv{u}_{sj}}(\bv{z}_{sj} > 0)\log\frac{\bs{\theta}_j}{1 - \bs{\theta}_j} + \log (1 - \bs{\theta}_j)\right), \label{eq:local_lb_ss_alt_smooth}
\end{align}
where $R_{\bv{u}_s}(\cdot)$ is the cumulative distribution function (CDF) of the continuous relaxation $r_{\bv{u}_s}(\cdot)$. We can now straightforwardly optimize the surrogate objective with gradient ascent.

\newpage
\section{\fedsparse{} algorithm}
\input{sections/fedsparse_algorithm}

\section{Alternative priors for the hierarchical model}
\label{sec:alt_priors_fl}
In the main text we argued that the hierarchical model interpretation is highly flexible and allows for straightforward extensions. In this section we will demonstrate two variants that use either a Laplace or a mixture of Gaussians prior, coupled with hard-EM for learning the parameters of the hierarchical model. 

\subsection{Federated learning with Laplace priors}
Lets start with the Laplace variant; the Laplace density for a specific local parameter $\bs{\phi}_{si}$ will be:
\begin{align}
    p(\bs{\phi}_{si} | \bv{w}_i) &= \frac{\lambda}{2} \exp(-\lambda |\bs{\phi}_{si} - \bv{w}_i|).
\end{align}
Therefore the local objective of each shard and the global objective will be
\begin{align}
    \mathcal{L}_s(\mathcal{D}_s, \bv{w}, \bv{w}_s) & := \sum_i^{N_s}L(\mathcal{D}_{si}, \bv{w}_s) - \lambda \sum_j |\bv{w}_{sj} - \bv{w}_j| + C,\\
    \mathcal{L}(\bv{w}) & := \sum_s \mathcal{L}_s(\mathcal{D}_s, \bv{w}, \bv{w}_s)
\end{align}
where again the global objective is a sum of all the local objectives and $C$ is a constant independent of the optimization. We can then proceed, in a similar fashion to traditional ``cross-device'' FL, by selecting a subset of shards $B$ to approximate the global objective. On these specific shards, we will then optimize $\mathcal{L}_s$ with respect to $\bv{w}_s$ while keeping $\bv{w}$ fixed. Interestingly, due to the $L_1$ regularization term that appears in the local objective, we will have, depending on the regularization strength $\lambda$, several local parameters $\bv{w}_s$ that will be exactly equal to the server parameters $\bv{w}$ even after optimization. Now given the optimized parameters from these shards, $\bv{w}_s^*$, we will update the server parameters $\bv{w}$ for the M-step by either a gradient update or a closed form solution. Taking the derivative of the global objective with respect to a $\bv{w}_j$ we see that it has the following simple form
\begin{align}
    \frac{\partial{L}}{\partial \bv{w_j}} = \lambda \sum_{s \in B} \text{sign}(\bv{w}_{sj} - \bv{w}_j).
\end{align}
By setting it to zero, we see that the closed form solution is again easy to obtain
\begin{align}
    \frac{\partial{L}}{\partial \bv{w}_j} = 0 \Rightarrow \lambda \sum_{s \in B} \text{sign}(\bv{w}_{sj} - \bv{w}_j) = 0 \Rightarrow \bv{w}_j = \text{median}(\bv{w}_{1j}, \dots, \bv{w}_{Bj})
\end{align}
since the median produces an equal number of positive and negative signs. Using the median in the server for updating its parameters is interesting, as it is more robust to ``outlier'' updates from the clients. 

\subsection{Federated learning with mixture of Gaussian priors}
The mixture of Gaussians prior will allow us to learn an ensemble of models at the server. The density for the entire vector of local parameters $\bv{w}_s$ in the case of $K$ equiprobable components in the mixture will will be
\begin{align}
    p(\bs{\phi}_s | \bv{w}_{1:K}) &= \frac{1}{K}\sum_k \mathcal{N}(\bv{w}_k, (1/\lambda) \bv{I}).
\end{align}
This will lead to the following local and global objectives 
\begin{align}
    \mathcal{L}_s(\mathcal{D}_s, \bv{w}_{1:K}, \bv{w}_s) & := \sum_i^{N_s}L(\mathcal{D}_{si}, \bv{w}_s) + \log \sum_k \frac{Z_k}{K} \exp\left(-\frac{\lambda}{2} \|\bv{w}_s - \bv{w}_k\|^2\right),\\
    \mathcal{L}(\bv{w}_{1:K}) & := \sum_s \mathcal{L}_s(\mathcal{D}_s, \bv{w}_{1:K}, \bv{w}_s)
\end{align}
where $Z_k$ is the normalizing constant of component $k$. Now we can proceed in a similar fashion and select a subset of shards $B$ to obtain the $\bs{\phi}_s^*$ while keeping the parameters of the prior $\bv{w}_{1:K}$ fixed. Now by taking the gradient with respect to one of the members of the ensemble $\bv{w}_k$, given the optimized local parameters $\bs{\phi}_s^*$, we see that
\begin{align}
    \frac{\partial L}{\partial \bv{w}_k} = \lambda \sum_s \frac{\frac{Z_k}{K} \exp(-\frac{\lambda}{2} \|\bv{w}_s - \bv{w}_k\|^2)}{\sum_j\frac{Z_j}{K} \exp(-\frac{\lambda}{2} \|\bv{w}_s - \bv{w}_j\|^2)} (\bv{w}_s - \bv{w}_k).
\end{align}
Notice that this gradient estimate can also be interpreted in terms of a posterior distribution over the index $z$ (taking values in $\{1, \dots, K\}$) given the ``observed'' variables $\bv{w}_s$; we can treat $1/K$ as the prior probability of selecting component $z = k$, i.e., $p(z = k)$ and $Z_k \exp(-\frac{\lambda}{2} \|\bv{w}_s - \bv{w}_k\|)$ as the probability of $\bv{w}_s$ under Gaussian component $k$. In this way, the weighting term can be written as
\begin{align}
    \frac{\frac{Z_k}{K} \exp(-\frac{\lambda}{2} \|\bv{w}_s - \bv{w}_k\|^2)}{\sum_j\frac{Z_j}{K} \exp(-\frac{\lambda}{2} \|\bv{w}_s - \bv{w}_j\|^2)} = p(z = k| \bv{w}_s, \bv{w}_{1:K}, \lambda).
\end{align}
This make apparent the connection to Gaussian mixture models, since $p(z = k| \bv{w}_s, \bv{w}_{1:K}, \lambda)$ is equivalent to the responsibility of component $k$ generating $\bv{w}_s$. Now we can again find a closed form solution for $\bv{w}_k$ by setting the derivative to zero
\begin{align}
    \frac{\partial L}{\partial \bv{w}_k} = 0 \Rightarrow \lambda \sum_s p(z=k|\bv{w}_s, \bv{w}_{1:K}, \lambda) (\bv{w}_s - \bv{w}_k) \Rightarrow \bv{w}_k = \sum_s \frac{p(z=k|\bv{w}_s, \bv{w}_{1:K}, \lambda)}{\sum_j p(z=k|\bv{w}_j, \bv{w}_{1:K}, \lambda)}\bv{w}_s,
\end{align}
which is again similar to the closed form update for the centroids in a Gaussian mixture model when trained with EM.

%% file: sections/fedsparse_algorithm.tex
\begin{algorithm}[h]
\caption{The server side algorithm for \fedsparse{} (assuming weight sparsity for simplicity). $\sigma(\cdot)$ is the sigmoid function, $\epsilon$ is the threshold for pruning.}\label{alg:fedsparse}
\begin{algorithmic}
\State Initialize $\bv{v}$ and $\bv{w}$ 
\For{round $t$ in $1, \dots T$}
    \State $\bs{\tau} \gets \log(1 + \exp(\bv{v}))$
    \State $\bs{\theta} \gets \sigma\left((|\bv{w}| - \bs{\tau}) / T\right)$
    \State $\bv{w} \gets \mathbb{I}[\bs{\theta} > \epsilon] \bv{w}$  \Comment{prune global model}
    \State Initialize $\nabla_{\bv{w}}^t = \mathbf{0}, \nabla_{\bv{v}}^t = \mathbf{0}$
    \For{$s$ in random subset of the clients}
        \State $\bv{z}_s, \hat{\bv{w}}^t_{s} \gets$ \Call{Client}{$s, \bv{w}, \bv{v}$}
        \State $\nabla_{\bv{w}}^t += \bv{z}_s (\hat{\bv{w}}^t_{s} - \bv{w})$
        \State $\nabla_{\bv{v}}^t += - \left(\bv{z}_s (1 - \bs{\theta}) - (1 - \bv{z}_s)\bs{\theta}\right) \sigma(\bv{v}) / T $
    \EndFor
    \State $\bv{w}^{t+1}, \bv{v}^{t+1} \gets$ \Call{Adam}{$\nabla_{\bv{w}}^t$}, \Call{Adamax}{$\nabla_{\bv{v}}^t$}
\EndFor
\end{algorithmic}
\end{algorithm}

\begin{algorithm}[h]
\caption{The client side algorithm for \fedsparse{}.}\label{alg:fedsparse2}
\begin{algorithmic}
\State Get $\bv{w}, \bv{v}$ from the server
\State $\bs{\theta} \gets \sigma\left((|\bv{w}| - \bs{\tau}) / T\right)$
\State $\bv{w}_s, \bv{v}_s \gets \bv{w}, \bv{v}$
\For{epoch $e$ in $1, \dots, E$}
    \For {batch $b \in B$}
        \State $\bs{\tau}_s \gets \log(1 + \exp(\bv{v}_s))$
        \State $\bs{\pi}_s \gets \sigma\left((|\bv{w}_s| - \bs{\tau}_s) / T\right)$
        \State $L_s \gets \mathcal{L}_s(b, \bv{w}, \bs{\theta}, \bv{w}_s, \bs{\pi}_s)$
        \State $\bv{w}_s \gets$ \Call{SGD}{$\nabla_{\bv{w}_s}L_s$}
        \State $\bv{v}_s \gets$ \Call{Adamax}{$\nabla_{\bv{v}_s}L_s$}
    \EndFor
\EndFor
\State $\bs{\pi}_s \gets \sigma\left((|\bv{w}_s| - \bs{\tau}_s) / T\right)$
\State $\bv{z}_s \sim q_{\bs{\pi}_s}(\bv{z}_s)$\\
\Return $\bv{z}_s$, $\bv{z}_s \odot \bv{w}_s$
\end{algorithmic}
\end{algorithm}